\documentclass[journal]{IEEEtran}

\usepackage{times}
\usepackage{epsfig}
\usepackage{graphicx}
\usepackage{amsmath}
\usepackage{amssymb}

\usepackage[frozencache,cachedir=.]{minted}

\usepackage[utf8]{inputenc}
\usepackage[english]{babel}
\usepackage[inline]{enumitem}
\usepackage{subcaption}
\ifCLASSINFOpdf
\else
\fi

\begin{document}
\title{Embedding Earth: Self-supervised contrastive pre-training for dense land cover classification}

% \author{Michail Tarasiou,~\IEEEmembership{Member,~IEEE,}
%         and~Stefanos Zafeiriou,~\IEEEmembership{Fellow,~IEEE}}
\author{Michail Tarasiou and Stefanos Zafeiriou}
% \thanks{M. Shell was with the Department
% of Electrical and Computer Engineering, Georgia Institute of Technology, Atlanta,
% GA, 30332 USA e-mail: (see http://www.michaelshell.org/contact.html).}% <-this % stops a space
% \thanks{J. Doe and J. Doe are with Anonymous University.}% <-this % stops a space
% \thanks{Manuscript received April 19, 2005; revised August 26, 2015.}}

\markboth{Journal of \LaTeX\ Class Files,~Vol.~14, No.~8, August~2015}%
{Shell \MakeLowercase{\textit{et al.}}: Bare Demo of IEEEtran.cls for IEEE Journals}

\maketitle

\begin{abstract}
In training machine learning models for land cover semantic segmentation there is a stark contrast between the availability of satellite imagery to be used as inputs and ground truth data to enable supervised learning. While thousands of new satellite images become freely available on a daily basis, getting ground truth data is still very challenging, time consuming and costly. 
In this paper we present {\it Embedding Earth} a self-supervised contrastive pre-training method for leveraging the large availability of satellite imagery to improve performance on downstream dense land cover classification tasks. % dense pairwise contrastive  learning at the level  of pixels
Performing an extensive experimental evaluation spanning four countries and two continents we use models pre-trained with our proposed method as initialization points for supervised land cover semantic segmentation and observe significant improvements up to $25\%$ absolute {\it mIoU}. In every case tested we outperform random initialization, especially so when ground truth data are scarse. Through a series of ablation studies we explore the qualities of the proposed approach and find that learnt features can generalize between disparate regions opening up the possibility of using the proposed pre-training scheme as a replacement to random initialization for Earth observation tasks. Code will be uploaded soon at \verb!https://github.com/michaeltrs/DeepSatModels!.  

% We find the proposed method to improve segmentation accuracies in every case tested. 
% In this paper we present  the first self-supervised method for land cover segmentation with time series of satellite images based on . 
% and a series of ablation studies 
% First land cover segmentation model to generalize across different regions

\end{abstract}

% \fi
\IEEEpeerreviewmaketitle

\section{Introduction}
% In   adopting   the   2030   Agenda   for   Sustainable Development,  world  leaders  agreed  that  a  global Indicator  framework  was  necessary  to  measure, monitor   and   report   progress   towards   the   17 transformational   Sustainable   Development   Goals (SDGs)  and  169  associated  Targets.  
% Need
In recent years an increasing number of organizations have placed emphasis on the use of Earth observation (EO) data and machine learning technologies as a means to partially automate or enchance their capacity in setting, monitoring and reporting the progress made on their associated goals and targets. Such technologies could be applied for projects ranging from global scale to benefiting targeted communities in need \cite{KANSAKAR201646}.
% pattern recognition
The United Nations (UN) aim at using remote sensing data for monitoring sustainable development goals \cite{UNdevgoals}. In the European Union (EU) part of the Common Agricultural Policy (CAP) objectives is the use of remote sensing tools for the automation and monitoring of CAP allocations and sustainability targets \footnote{http://esa-sen4cap.org/content/agricultural-practices} \footnote{http://esa-sen4cap.org/content/crop-diversification}. In Africa the use of satellite imagery for crop yield prediction and early detection of pests and plant diseases could lead to almost real time monitoring of food supplies and the prevention of potential shortages \cite{MUTANGA2017231}. 
% climate change, crop health for food security, Accurate early detection and diagnosis of crop diseases and pests is key factor in plant production at both farm and local scale.

% Capacity
At the same time driven by potential benefits an increasing number of EO satellites are getting launched collecting data for Earth systems and monitoring the variability in natural and built environments. 
% Currently there are more than 150 such constellations in orbit constantly transmitting data back to Earth. 
For Sentinel-1/2 satellite missions alone the total amount of data available to download has already reached the Petabyte scale. 
% improving decision making and efficient resource allocation
% single image (blue marble) credited with the rise of th environmental movement, perhaps an earth observation database will lead to the next steps in managing the fragile ecosystem of our planet
% Problem - bottleneck
This plethora of input data is in stark contrast to the difficulty of obtaining ground truth annotations, especially at the scale required to make the most out of supervised deep learning alghorithms which have revolutionized computer vision \cite{lenet, alexnet, resnet}. Compiling ground truth datasets for EO tasks involves the collection of data either through the deployment of field missions or the development of platforms for crowd sourcing data \cite{crowdsource}. Both solutions are time consuming and costly especially so for remote and difficult to access regions. % The amount of ground truth data available is also highly non uniform and depends an emerging pattern of data rich vs poor regions. In Europe it is possible to compile large datasets of . However, for other regions data can be really sparse. 

% Solution
This imbalance necessitates the development of techniques for learning without annotations that can leverage the large amounts of readily available available EO data. % to increase performance in the absence of large scale annotations. 
% pretrain then finetune in downstream task framework
In this paper, we propose such a method for self-supervised learning of dense representations. Our key contributions are the following:
\begin{enumerate}
    \item we design a system for end-to-end self-supervised visual pre-training at the pixel level. Finetuning the pre-trained models on land cover semantic segmentation we observe significant improvements up to $25\%$ absolute {\it mIoU} compared to our baselines. Benefits are particularly pronounced for datasets containing relatively few annotations.
    \item through a series of ablation studies we explore the method hyperparameter space and derive conclusions about how the size and qualitative characteristics of the pre-training set affect downstream segmentation performance. We find that increasing the size of the pre-training set, either by extending the {\it Period of Interest} (POI) or the {\it Area of Interest} (AOI), generally improves downstream segmentation performance. 
    \item we find no performance deterioration when initializing from a model pre-trained in a different region than the AOI of the downstream segmentation task, e.g. pre-training a model on a different country compared to pre-training using at the same AOI as the annotated dataset. This observation opens the possibility for the general use of such pre-trained models as initialization points for land cover segmentation tasks. We make the pre-trained models used in this study publicly available together with code for replicating the experiments presented here.   
\end{enumerate}

%The employed pretext task is not particularly tailored to land cover classification but is general enough for the method to offer .\\

% high resolution multispectral imagery

\section{Related work}
\subsection{Beyond random initialization of model parameters} 
When training DNNs an unfavourable initialization of model parameters can lead to several problems such as parameter saturation and very slow convergence to a good solution \cite{glorot_init}. Several works have proposed techniques for stabilizing the training dynamics making it possible to train deeper architectures that lead to improved DNN performance overall \cite{glorot_init, he_init, init1, init2}. %leading to capacity to train deeper networks than it would be possible with basic random initialization 
Beyond these insights regarding parameter initialization it has been found that rather than randomly initializing network parameters for every new task, using the weights of a model previously trained on a different domain can in many cases improve performance. {\it Transfer Learning} revolves around this idea: how to transfer knowledge which has been learnt into one {\it source} domain to another downstream {\it target} domain. A common practice for training DNNs for vision related tasks has been to use models trained on large scale image classification datasets, e.g. {\it ImageNet} \cite{imagenet} as an initialization point for downstream tasks \cite{imagenet_pretrain}. For this setting to work knowledge learnt during pre-training needs to be general enough to find some application in the target domain. In that spirit it is convenient if there is some overlap between the semantic objects found in the pre-training and target domain datasets. %\cite{rethink_imagenet}
Another approach following similar principles is to use un/self-supervised training prior to fine-tuning a network into the target domain. This approach has also been shown to improve performance \cite{erhan} and has two distinct advantages compared to supervised transfer learning. First, unlabelled data are inexpensive to obtain making it easier to compile large scale datasets which contain significant variation and can be used to train models that generalize well to unseen data.% and can lead to knowledge during pre-training. 
Secondly, it makes it possible to compile a dataset and derive a pre-training methodology that is specifically tailored to the target domain and task. 
For dense classification tasks with remote sensing imagery our proposed method follows the latter approach of self-supervised pre-training. The main reason for this choice is the difficulty of obtaining annotations at high resolution over very large AOI and the inability of land cover classification models to generalize from one region to another as has been empirically shown in previous studies \cite{generalize1, breizhcrops2020}.   
% We distinguish between two different approaches commonly used to enhance performance w.r.t. training a model starting from randomly initialized parameters, namely and  

\subsection{Metric learning}
The main goal of metric learning is to express the similarity between different inputs by use of a distance metric  over their encoded representations. 
For training DNNs the most commonly encountered framework for applying metric learning is to use {\it Siamese Networks} trained with a {\it Contrastive} loss function. {\it Siamese Networks} employ two copies of the same network used to process two inputs of known semantic similarity and obtain their respective encodings. Popular applications involve signature \cite{siam_signature} and face \cite{siam_face} verification and one-shot image recognition \cite{siam_oneshot}. 
A contrastive loss function is designed for reducing the distance between encodings of inputs characterized by similar semantics (positive pairs) and increasing the distance between pairs with different semantics (negative pairs). Originally introduced for face verification \cite{siam_face} there has recently been a resurgence of {\it Contrastive} learning methods in computer vision as a means for learning a useful embedding space and to enchance the discriminatory capacity of DNNs on downstream tasks. The method can be applied in either a fully supervised manner \cite{contr_sup1, contr_sup2, Chung16a, triplet, cscl} or, by choice of a suitable pretext task, as a self-supervised learning task \cite{inst_discr, cpc, simclr}. %In this manner the network learns an embedding space mapping similar samples close to one another and dissimilar samples far

% Siamese networks [4] are general mod-els for comparing entities.
% The core idea of contrastive learn-ing [16] is to attract the positive sample pairs and repulse thenegative sample pairs. This methodology has been recentlypopularized for un-/self-supervised representation learning[36, 30, 20, 37, 21, 2, 35, 17, 29, 8, 9]. Simple and effectiveinstantiations of contrastive learning have been developedusing Siamese networks [37, 2, 17, 8, 9].

% lecun
% what is pretext task
\subsection{Self-supervised learning for vision tasks}
Before computer vision self-supervised learning approaches have driven a revolution in natural language processing (NLP). Models pre-trained on a large corpus of text significantly outperform their counterparts trained only using annotated data \cite{bert, roberta, bigbird}. These methods typically employ autoencoders \cite{autoencoders, autoencoders1, autoencoders2, autoencoders3} and in particular the masked autoencoder framework \cite{made} in which part of the input is hidden and the learning task is to predict that in the output. For successfully doing so the model is expected to learn some underlying structure of the data and thus learn about the data space without requiring annotations.  
% Many approaches to learn useful features in an  unsupervised learning setting can be found in literature. Here we present relate works from two distinctive approaches which either learn to reconstruct part of the input or to discriminate every instance in a training set.\\
% The former category has a simple objective which is to hide part of the input data from the model and learn to predict what was hidden in an effort to learn some underlying structure of the data. 
Several works in computer vision have used the same framework. Applying the {\it BERT} criterion \cite{bert} on natural images \cite{bert_imnet} hide and subsequently predict a fixed ratio of pixels in an autoregressive manner. {\it Contrastive Predictive Coding} (CPC) \cite{cpc, cpc2} uses autoregressive modelling to predict the future in latent rather than input space. While this framework naturally suits problems involving sequential data a variety of problems can be formulated using this framework, e.g. for image classification an image can be seen as a series of patches extracted from a predefined grid.\\ 
A recently popularized framework for self-supervised learning in vision involves using instance discrimination \cite{inst_discr} as a pretext task. Instance discrimination assumes that each sample in a training set belongs to a class of each own and the training task involves learning to discriminate each sample from all other samples. In a contrastive learning setting two different views of the same sample, obtained by data augmentation, should be mapped nearby in embedding space while all other inputs should be contrasted. We note how this framework does not require any annotations, the identities of the training samples alone are used to define a set of classes. 
In that spirit \cite{inst_discr, simclr, simclrv2, moco, mocov2} use the instance discrimination task and variants of {\it Noise Contrastive Estimation} (NCE) loss \cite{nce}.  %to discriminate each sample individually learning 
They manage to learn high quality embedding spaces judged by the capacity of a linear classifier to separate produced embeddings into classes when trained on annotated data. These methods have contributed to closing the gap in the discriminative power between models trained in a fully supervised and self-supervised manner. 
It is generally agreed that comparing with a large number of negative pairs is required for a successful implementation. In \cite{simclr, simclrv2} this is achieved by use of a very large batch size. However, this imposes a requirement for large accelerator memory. To decouple the number of negative samples from the mini-batch size, allowing it to be large, \cite{inst_discr} keep encodings for every sample in the training set in a memory bank which is updated when the same sample is encountered during each epoch. In the same spirit \cite{moco, mocov2} maintain a dynamic dictionary of encodings as a fixed size data queue. Our method uses the latter approach which is found to improve performance without the need of large batch sizes. Also, strong data augmentation and the addition of a separate 2-layer predictor network for obtaining final embeddings were both shown to improve performance in \cite{simclr, mocov2}. In our proposed approach which is designed for use with satellite imagery data augmentation is driven by the characteristics of input domain. Additionally in preliminary experiments the use of a predictor network was not found to improve performance and is avoided in favour of simplicity.

% A recent body of works has 

All the above works were developed to learn global image representations suitable for image classification. More recently self-supervised instance discrimination was extended to a dense context in \cite{pinheiro, wang}. Both works employ data queues and input augmentation to create two different views of a sample image. Positive pairs are obtained by sampling from a set of pixels in correspondence between the two views. While \cite{pinheiro} keep track of augmentation parameters to derive pixel correspondences \cite{wang} opt to learn correspondences as a separate process during training. Additionally, \cite{wang} add a loss component on global features which they show to stabilize the learning process. Our work is more similar to \cite{pinheiro} in that we only use local features and calculate explicit pixel correspondences. In contrast to both works we formulate our loss function to include pixels from the same sample into the set of negative pairs and extend the data augmentation module for use with image time-series.

\subsection{Land cover semantic segmentation using satellite imagery}
Land cover semantic segmentation aims at classifying each location found in a satellite image, or a set of images, as one out of a predefined number of land cover types. The same techniques for training DNNs that have lead to significant advances in computer vision community have been consistently shown to outperform previous remote sensing based approaches either using single images \cite{agrvision_challenge, agrvision_results} or image time series of \cite{Ruwurm1, mtlcc, Rustowicz2019SemanticSO, garnot2019satellite}. 
However, the domain of satellite images is fundamentally different from that of natural images requiring special treatment for successful model training. For example while it is possible to compile large scale datasets of natural images from a variety of image sensors this is not possible with remote sensing data as satellite sensors are expensive to set in orbit and exhibit more variation between one another than RGB sensors do, e.g. in terms of resolution and image bands. As such most trained models tend to be sensor dependent. Moreover, regional variations in atmospheric conditions, land geometry and land cover types hinder the generalization capability of models trained with full supervision \cite{generalize1, breizhcrops2020} and obtaining large scale ground truth annotations at a scale that would make generalization possible has not been achieved thus far.\\
%Several studies have explored the use of DNNs for dense classification of remote sensing data either from static inputs \cite{} or from time series of satellite images \cite{Ruwurm1, mtlcc, Rustowicz2019SemanticSO, bert_sat}. Our work is more falls in th elatter category as we make use of time series objects.
Tackling semantic segmentation of crop types from time series of satellite images \cite{mtlcc} exploit the spatial temporal nature of the inputs and propose the use of {\it CRNN} framework \cite{conv_lstm} in which fully connected {\it RNN} layers are replaced by 2D convolutional layers. Improving their results \cite{Rustowicz2019SemanticSO} first process each image individually by a 2D {\it CNN} before applying a {\it CRNN} backend and also test a fully convolutional {\it 3D-UNET} \cite{3dunet}. Both models achieve similar performance. Exploring the use of pre-training methods \cite{cscl} use a fully supervised contrastive criterion to model the relationship between each pixel and its local spatial context. Their work shows improvements over the aforementioned studies \cite{mtlcc, Rustowicz2019SemanticSO} while exhibiting significant performance gains along parcel boundaries. 
Using \cite{domain_adapt1} use the BDL framework \cite{bdl} for semantic segmentation adversarial domain adaptation to improve the performance between static RGB images obtained from different satellites. 
In the self-supervised space \cite{bert_sat} use the {\it BERT} criterion on time series of satellite images improving the performance of randomly initialized baselines. Their approach only uses individual pixels and thus does not perform joint spatio-temporal modelling which is a goal of this study. For static images \cite{semantic_inpainting_sat} use semantic inpainting \cite{semantic_inpainting} for overhead scene parsing, road network extraction, and land cover estimation.
%In our work we use the {\it UNet3Df} model from \cite{cscl} as the backbone architecture for our experiments. Our method is based on the {\it Instance Discrimination} pretext task and requires no annotations. The inputs to our models are {\it Sentinel-2} satellite image time series.    

\begin{figure*}[!h]
\begin{center}
   \includegraphics[width=0.65\linewidth]{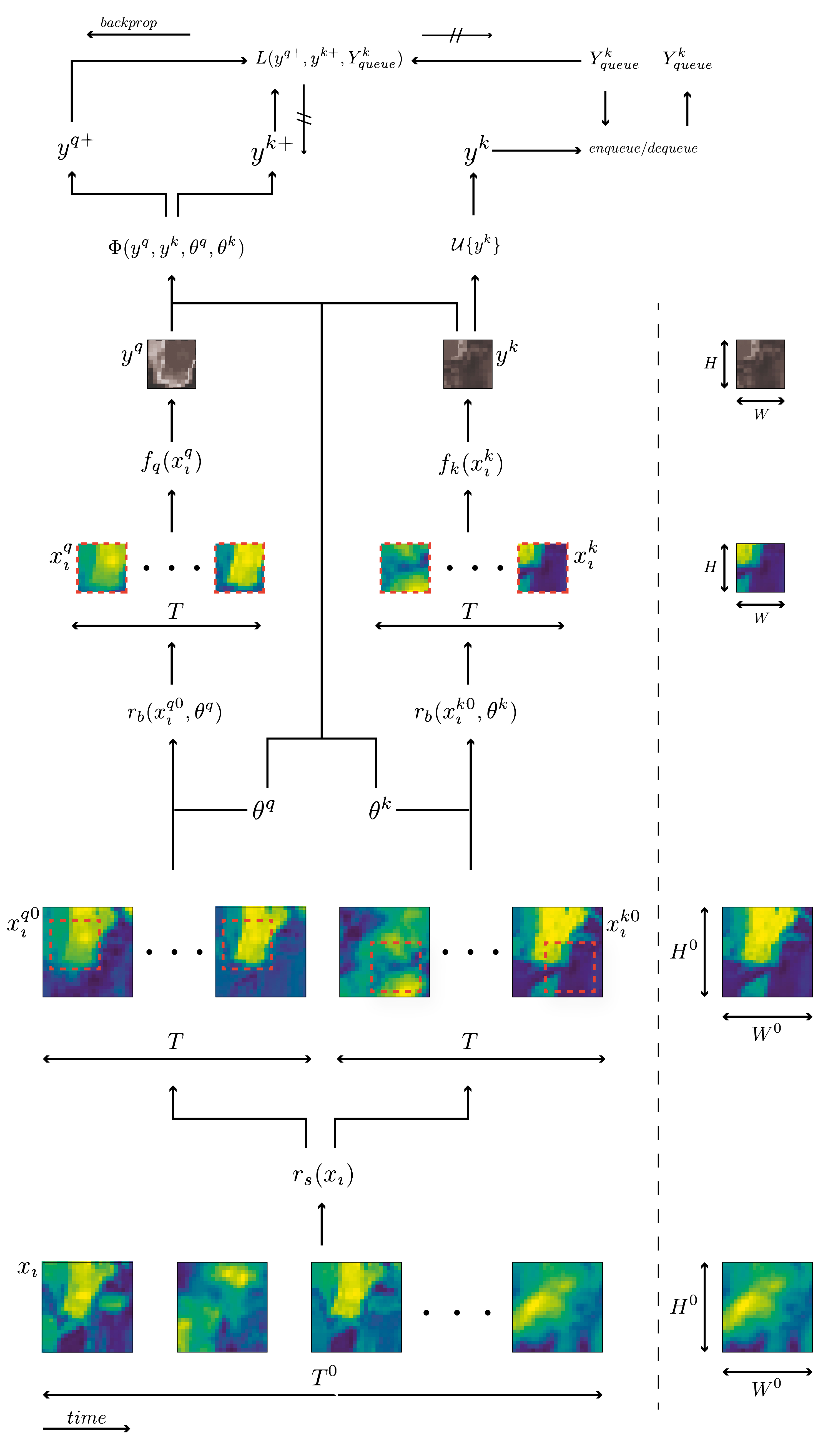}
\end{center}
   \caption{Overview of the proposed pre-training framework. Bottom to top: An input time series is processed by the sample ($r_s$) and batch ($r_b$) level augmentation modules to produce the inputs for the queries ($f_q$) and keys ($f_k$) encoders. The correspondence module $\Phi$ retrieves pixels in correspondence $y^{q+}$ and $y^{k+}$ from extracted embeddings $y^q$ and $y^k$ which are used to form positive and in-sample negative pairs. These features, together with the data queue $Y^k_{queue}$ are used to derive the current loss. Backpropagation is applied only for the queries encoder. After parameter update the earliest encodings are dequeued and $N_{queue}$ randomly sampled embeddings $y^k$ are enqueued.}
\label{fig:model}
\end{figure*}

\section{Proposed approach}\label{method}
% forcing localfeatures to remain constant over different viewing conditions (pin)
Our proposed method can be categorized as a dense instance discrimination pretext task. % similar to \cite{pinheiro, wang}. 
In particular we make the assumption that pixels corresponding to the same location over the span of a POI correspond to the same land cover type. Thus, we force extracted features corresponding to the same location under different views in space and time to be similar. At the same time we assume that no two locations have the exact same characteristics, thus, we optimize our feature extractor such that features extracted at different locations are dissimilar irrespective of view.
% pixel-level contrastive learning, in  embedding space
% finetuning (use pre-training to find a "better than random" initialization point) vs linear evaluation (used to estimate feature quality independently of model architecture, captures the intuition that high quality features can be used to linearly separate the classes found in the inputs)
Overall our training pipeline includes the following components:
\begin{enumerate}
    \item a stochastic data augmentation module $r(\cdot)$ is used to extract two different views for every sample in a mini-batch of image time series.
    \item queries and keys encoder functions $f_q(\cdot), f_k(\cdot)$ extract pixel level encodings for each respective view.
    \item a dense correspondence module $\Phi(\cdot)$ is used to find pairs of pixels corresponding to the same location from the two views (positive pairs). All other pairs are considered to be negative pairs,
    \item a data queue $K$ in which key encodings from the current mini-batch are enqueued while these from the oldest one are dequeued.
    \item a contrastive loss function $L(\cdot)$ is used to encourage extracted encodings from positive pairs to be close in embedding space and far for all negative pairs.   
\end{enumerate}
All model components are presented schematically in Fig.\ref{fig:model} and in Python pseudocode in Listing \ref{lst:the-code}. In the following sections we describe each component individually.
% feature invariance for the same location but different spatial context (different image crop) and sampled at different times during the year.feature invariance for the same location but different spatial context (different image crop) and sampled at different times during the year.
\subsection{Extracting different views}\label{view_extr}
% A stochasticdata augmentationmodule that transformsany given data example randomly resulting in two cor-related views of the same example, denoted ̃xiand ̃xj,which we consider as a positive pair.  In this work, wesequentially apply three simple augmentations:randomcroppingfollowed by resize back to the original size,ran-dom color distortions, andrandom Gaussian blur.  Asshown in Section 3, the combination of random crop andcolor distortion is crucial to achieve a good performance.
To ensure an efficient implementation our data augmentation module is decomposed into two separate modules $r=r_b \circ r_s$ at the sample and batch levels. Separately for each sample $x_i \in \mathbb{R}^{T_i \times H \times W \times C}$ in a mini-batch $X = \{x_i\}, i \in \{1,...,N\}$ the sample level module extracts two views $x_i^{qs}, x_i^{ks} = r_s(x_i) \in \mathbb{R}^{T \times H \times W \times C}, T < T_i$ sampling randomly in time. Outputs are concatenated into two separate mini-batches. For each of the two mini-batches the batch level module $r_b$ uses random augmentation parameters $\phi^q, \phi^k$ to extract two random crops $x_i^{q} = r_b(x_i^{qs}, \phi^k), x_i^{k} = r_b(x_i^{ks}, \phi^k) \in \mathbb{R}^{T \times H_{cr} \times W_{cr} \times C}$ while at the same time performs random horizontal and vertical flipping. In practice the augmentation parameters include the top-left coordinates of the crop, the crop size and whether the image was flipped in either dimension. The probability of flipping is set to $0.5$ in each dimension. The choice of crop size $H_{cr} \times W_{cr}$ w.r.t. the initial size of the images $H \times W$ is made such that there is a minimum overlap area $A_{min}$ between the two crops to facilitate the extraction of positive pairs for training as described next. 
% $b=\frac{1+\sqrt{r}a}{2}$ such that there is a minimum overlap ratio $r=\frac{A_{overlap}}{A}$, $A=a^2, A_{overlap}=c^2$ between crops as shown in Fig.\ref{fig:crops}.

% \begin{figure}[t]
% \begin{center}
%     % \fbox{\rule{0pt}{2in} \rule{0.9\linewidth}{0pt}}
%   \includegraphics[width=0.5\linewidth]{figures/crops.png}
% \end{center}
%   \caption{Data augmentation minimum overlap between image crops}
% \label{fig:crops}
% \end{figure}

%  \in \mathbb{R}^{N \times T \times H_0 \times W_0 \times C}
% $X^q_0, X^k_0$
\subsection{Feature extraction}
For a backbone $g: \mathbb{R}^{T \times H_{cr} \times W_{cr} \times C} \rightarrow \mathbb{R}^{H_{cr} \times W_{cr} \times D}$ extracted dense features are further processed by a projection head $h: \mathbb{R}^{H_{cr} \times W_{cr} \times D} \rightarrow \mathbb{R}^{H_{cr} \times W_{cr} \times D}$ which consists of a $1\times 1$ convolutional layer followed by a {\it ReLu} activation followed by a final $1\times 1$ convolutional layer. The complete network $f=h \circ g$ is initialized as queries encoder $f_q$ and its parameters are copied to produce the initialization point for the keys encoder $f_k$. Both are used to obtain final dense embeddings $y^q_i = f_q(x_i^q), y^k_i = f_k(x_i^k) \in \mathbb{R}^{H_{cr} \times W_{cr} \times D}$. After pre-training the projection head is discarded and a linear layer is added to project the final embedding to a space with the appropriate dimensionality. % as dictated by the total number of classes. 
% of choice $f$ mapping from the data input space $\mathbb{R}^{T \times H_{cr} \times W_{cr} \times C}$ to a $D$-dimensional embedding space $\mathbb{R}^{H_{cr} \times W_{cr} \times D}$  
%$f: \mathbb{R}^{T \times H_{cr} \times W_{cr} \times C} \rightarrow \mathbb{R}^{H_{cr} \times W_{cr} \times D}$ 
% we use random parameter initialization to begin training. 

\subsection{Positive and negative pair matching}
Positive pairs from the queries and keys encoders are pixels in correspondence between the two views. To find them we keep track of applied data augmentation parameters $\phi^q, \phi^k$ during each training step which we feed to a correspondence module $\Phi$ to get $N^+$ randomly sampled positive pairs $q_i, k_i = \Phi(y^q, y^k, \phi^q, \phi^k, N^+) \in \mathbb{R}^{N^+ \times D}$ from the total number of pixels in correspondence. $N^+$ is a hyperparameter which defines the number of positive pairs per training sample.
% pends on the number of pixels in correspondence $A \geq A_{min}$. This varies for each batch and is a function of the choice of crop and original image dimensions as described in section \ref{view_extr}. 
%From these pixels we sample $N$ locations randomly per data sample which are used as in-sample queries and keys $q=q^{s}, k^{s} \in \mathbb{R}^{N \times d}$ whose elements are in correspondence. 
Furthermore, we randomly sample another $N^{queue}$ locations from the encoded keys to obtain the negative pairs $k_i^{queue} \in \mathbb{R}^{N^{queue} \times D}$ which are enqueued in our keys data queue $K \in \mathbb{R}^{M \times d}$. Enqueing and dequeuing $K$ takes place after weights are updated during each training step to avoid to possibility to to include positive pairs into the data queue.  
% $q, k \in \mathbb{R}^d$
%we opt for deriving pixel correspondences from known data augmentation and take the additional step of controlling the locations of positive and negative features to allow even sampling along spatial dimensions (?). However, we use mean features for all negative samples added to MoCo queue similar to \cite{wang}. 

\subsection{Loss function}
To learn an instance discriminative feature space we need a loss function that encourages the each encoded query $q_j$ to be similar with its corresponding key $k_j$ and dissimilar with all other in in-sample keys $k_i, i \neq j$ and queue keys $K_i$. For this purpose we use the loss function from \cite{inst_discr, cpc} which takes the following form for our pretext task.

\begin{equation}\label{losseq}
L = - \sum_{j=1}^N \log \frac{\exp(q_j k_j / \tau)}{\sum_{i=1}^{N} \exp(q_j k_i  / \tau) + \sum_{i=1}^{M} \exp(q_j K_i  / \tau)}    
\end{equation}
where $\tau$ is a temperature hyperparameter that controls the concentration of the probability distribution imposed by the {\it SoftMax} layer \cite{distil, contr_temperature} and $M$ is the size of our data queue. The similarity between encodings is measured by their dot-product. We note that all elements of $q, k, K$ are normalized by their $L2$-norm prior to application of eq.\ref{losseq}. It is interesting to view eq.\ref{losseq} as a non-parametric version of the {\it Cross-Entropy Loss} function with a {\it SoftMax} classifier in which $k_j$ is the true class of $q_j$. 

% comment on temperature
% As the temperatureτdecreases, the entropy of the distributionH(ri)decreasesstrictly (the proof is in supplementary material).  The dis-tribution ofribecomes more sharp on the large similarityregion,  which  gives  large  penalties  to  the  samples  closed to x_i

\subsection{Parameter update}
% More-over, as the dictionary keys come from the preceding sev-eral mini-batches, aslowly progressingkey encoder, imple-mented as a momentum-based moving average of the queryencoder, is proposed to maintain consistency
While the {\it MoCo} mechanism proposed in \cite{moco} effectively decouples the number of negative samples from mini-batch size it also makes it impossible to update the parameters of the keys encoder via back-propagation as the hidden layer features are discarded to reduce memory requirements. An obvious choice would be to only use a single encoder network and use it for getting both queries, keys and to update the queue values. However, in \cite{moco} it was noted that this approach yields poor performance in practice. This is attributed to the fact that a fast parameter update of the encoder practically renders earlier queue elements inconsistent with the current mini-batch for similarity training. 
To counter both problems they propose the following methodology which we also employ in this paper. Both queries and keys encoders are initialized with the same parameters. At each training step only the parameters of the queries encoder are updated by back-propagation while the parameters of the keys encoder are updated as an exponential moving average of the queries encoder.
\begin{equation}
    \theta_q \xleftarrow{} \theta_q - \alpha \nabla_{\theta_q} L 
\end{equation}

\begin{equation}
    \theta_k \xleftarrow{} m \theta_k + (1-m)\theta_q
\end{equation}
Where $\theta_q, \theta_k$ are the parameters of the queries and keys encoders, $\alpha$ is the learning rate and $m \in [0, 1)$ is the momentum coefficient.
This way it is possible to differentiate $f_q$ from $f_k$ while at the same time $f_k$ evolves much slower allowing earlier batches maintain their contribution to the loss function. %In this manner making it possible to use a large queue while keys fro. %In fact it is observed that large values for the momentum parameter work much better than smaller ones
\begin{listing}[!h]
\begin{minted}{python}
# f_q, f_k: encoder networks for query and key
# r: data augmentation module
# queue: queue of K keys (KxD)
# m: momentum
# t: temperature
f_k.params = f_q.params # initialize
for x in loader: # load a minibatch of B samples
  # augment and retain augmentation information
  x^q, aug^q = r(x) # BxTxHxWxC
  x^k, aug^k = r(x) # BxTxHxWxC
  # extract queries and keys
  q = norm(f_q(x^q)) # BxHxWxD
  k = norm(f_k(x^k)) # BxHxWxD
  k = k.detach() # no gradient to keys
  # rand sample N pixels/sample in correspondence 
  q, k^s = samplec(q, k, aug_q, aug_k, N) # BNxD
  # in-sample positive and negative logits
  l_ins = einsum('nc,kc->nk', [q, k^s]) # BNxBN
  # queue negative logits 
  l_neg = einsum('nc,kc->nk', [q, queue]) # BNxK
  # concat logits: 
  logits = cat([l_pos, l_neg], dim=1) # BNx(N+K)
  # positives are diagonal of l_ins
  labels = range(B*N)  
  # contrastive loss eq.1
  loss = CrossEntropyLoss(logits/t, labels)
  # SGD update for query network
  loss.backward()
  # momentum update for key network
  f_k.params = m*f_k.params+(1-m)*f_q.params  
  # randomly sample M queue elements per sample
  k^q = sample(k, M) # BMxD
  # update queue for current minibatch
  enqueue_and_dequeue(queue, k^q) 
\end{minted}    
    \caption{Pseudocode of proposed pre-train task in a PyTorch-like style.}  
    \label{lst:the-code} 
\end{listing}
% \subsection{Implementation details}
% For our experiments we use $N=M=32$, although different number of samples has been used successfully. $K$ and batch size. We refer the reader to \cite{moco} for further details
% shuffling batch norm
% do not rescale image. since learning for a single satellite/sensor we want to retain the same pixel size (in land m)

\section{Experiments}
% we compare our method to N different types of baselines: random initialization, supervised segmentation training and supervised pre-training on a pretext task in the same region \cite{cscl}.
% for Ghana and S.Sudan - explain why some runms are missing e.g. GHA-space
In all experiments presented our aim is to test the benefit of the proposed pre-training method as an alternative to random initialization for dense land cover classification. All experiments take the following form:
\begin{enumerate}
    \item a baseline performance for a model of choice and dataset is obtained by randomly initializing the network parameters and training with full supervision
    \item the same network architecture is trained with the methodology presented in section \ref{method} starting from random initialization using a large amount of unlabelled data ($N_{pretrain} \geq N_{train}$). The pre-trained model is then used as the initialization point for fully supervised end-to-end training using the same protocol as in step 1.
\end{enumerate}
For each experiment we denote both the pre-training and training datasets. The names of the training sets include a country code and a number indicating the logarithm of the subsampling ratio w.r.t. the total number of samples available for each respective country. For example dataset DEU0 includes all training samples in Germany (subsampled at a ratio of $2^{-0}=1$) while dataset GHA5 includes $2^{-5}=\frac{1}{32}$ of the number of data available in Ghana. % To test the quality of unsupervised features learnt during pre-training we are following two finetuning protocols, either training only a linear classifier while fixing the feature extractor weights or finetuning the entire network in an end-to-end manner. 
For all datasets we perform an ablation study by varying the amount of available annotations during training. The evaluation set is always kept constant between runs for the same country. As such there are numerous cases in which the number of evaluation data is much larger than the number of data used in training. In this section we present {\it mIoU} plots for a quick visual interpretation of results. Detailed performance metrics can be found in Appendix \ref{appendix_eval_metrics}.% An overview of the datasets used can be seen in Table \ref{datasets}. %Thus, we  of cheap unlabelled 
%{\bf Dataset size}. Pretraining on a fixed size of "cheap" unlabelled data and testing the performance by varying the amount of "expensive" fully annotated data. Expect to see larger differences for small annotations. Possibly perform the same experiment the other way around, varying unlabelled data and test on fixed small size annotations. In both cases annotated data are spread uniformly in the AOI.\\
%{\bf Dataset spread}. Train with localized data and test with spread eval data. Expect to see larger differences than above
%{\bf Pretraining with agricultural only vs all regions}. Test whether filtering out and selecting only agricultural regions for pretraining has any benefit over just extracting regions.\\
%{\bf Spatial proximity to AOI}. Test whether picking data nearby the AOI has any benefit for the downstream task.\\
%{\bf Yield crop prediction} should benefit as well.\\

\begin{table}[!ht]
\begin{center}
\begin{tabular}{|c|c|c|c|c|c|}
\hline
Name & $N_t$ & $N_y$ & $N_s$ & $D_{sample}$ & $D_{crop}$ \\
\hline \hline
FRA-base & 1 & 1 & 51,984 & 48X48 & 32X32\\
DEU-base & 1 & 1 & 117,649 & 32X32 & 24X24\\
GHA-base & 1 & 1 & 11,733 & 100X100 & 64X64\\
SSU-base & 1 & 1 & 11,881 & 100X100 & 64X64\\
\hline
FRA-space & 3 & 1 & 150,232 & 48X48 & 32X32\\
%DEU-space & DEU & 3 & 1 & & 32\\
%GHA-space & GHA & 1 & 3 & & 100\\
% SSU-space & 3 & 1 & 35,643 & 100X100 & 64X64\\
\hline
FRA-time & 1 & 3 & 152,438 & 48X48 & 32X32\\
DEU-time & 1 & 3 & 351,947 & 32X32 & 24X24\\
%GHA-time & GHA & 1 & 3 & & 100\\
SSU-time & 1 & 3 & 35,643 & 100X100 & 64X64\\
\hline
\end{tabular}
\end{center}
\caption{Pretraining datasets. $N_t$: number of Sentinel-2 tiles, $N_y$: number of years, $N_s$: number of samples, $D_{sample}$: resolution of images before data augmentation, $D_{crop}$: resolution of training images after data augmentation}
\label{unsup_datasets}
\end{table}

\begin{table}[!ht]
\begin{center}
\begin{tabular}{|c|c|c|c|c|}
\hline
Name & $N_y$ & $N_s$ & $N_{cl}$ & $D_{sample}$\\
\hline \hline
FRA-train &1 & 26,106 & 13 & 48X48\\
FRA-eval & 1 & 4,098 & 13 & 48X48\\
DEU-train & 1 & 27,053 & 17 & 24X24\\
DEU-eval & 1 & 8,391 & 17 & 24X24\\
GHA-train & 1 & 2,423 & 5 & 64X64\\
GHA-eval & 1 & 1,616 & 5 & 64X64\\
SSU-train & 1 & 3,510 & 6 & 64X64\\
SSU-eval & 1 & 2,344 & 6 & 64X64\\
\hline
\end{tabular}
\end{center}
\caption{Training datasets. $N_y$: number of years, $N_s$: number of samples, $N_{cl}$: number of classes, $D_{sample}$: resolution of images before data augmentation}
\label{sup datasets}
\end{table}

\subsection{Datasets}

\begin{figure*}[!tbp]
  \centering
  \begin{minipage}[b]{0.475\textwidth}
    \centering
    \includegraphics[width=1.\textwidth, trim={0 10 0 0}]{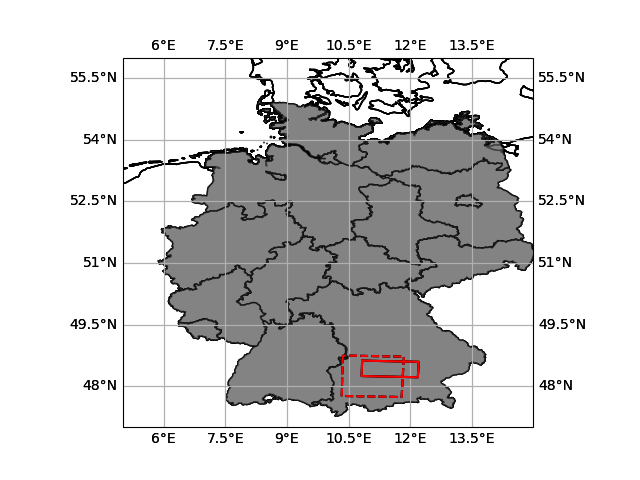}
    \caption{AOI in Germany. Solid lines denote the annotated AOI while dashed lines denote the area covered by DEU-base and DEU-time pre-training data.} %DEU-base and DEU-time cover the same AOI.}
    \label{map_DEU}
  \end{minipage}
  \hfill
  \begin{minipage}[b]{0.475\textwidth}
  % trim={<left> <lower> <right> <upper>}
    \centering
    \includegraphics[trim={0 30 0 0}, width=0.85\textwidth]{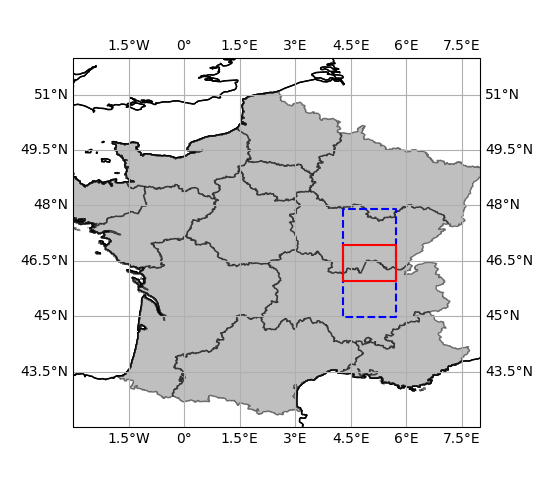}
    \caption{AOI in France. Solid lines denote the annotated AOI while dashed blue lines denote the area covered by FRA-space pre-training data. FRA-base and FRA-time datasets cover the same AOI as the annotated dataset.}
    \label{map_FRA}
  \end{minipage}
\end{figure*}

\begin{figure*}[!tbp]
  \centering
  \begin{minipage}[b]{0.475\textwidth}
    \centering
    \includegraphics[width=0.6\textwidth, trim={7 3 7 2}]{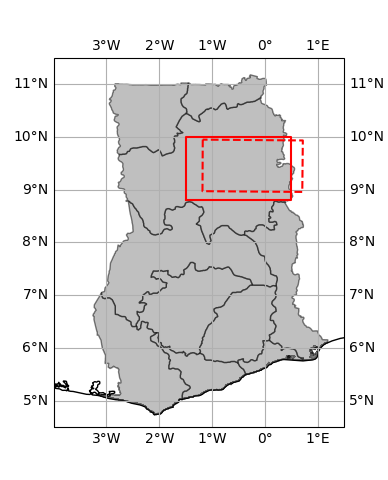}
    \caption{AOI in Ghana. Solid lines denote the annotated AOI while dashed lines denote the area covered by GHA-base pre-training data.}
    \label{map_GHA}
  \end{minipage}
  \hfill
  \begin{minipage}[b]{0.475\textwidth}
  % trim={<left> <lower> <right> <upper>}
    \centering
    \includegraphics[trim={0 0 0 0}, width=\textwidth]{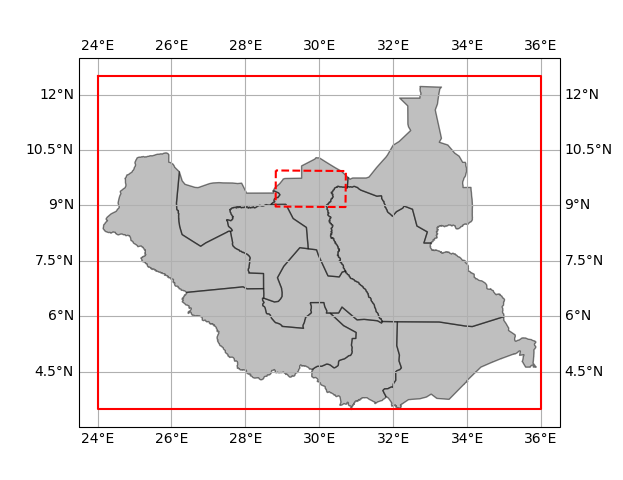}
    \caption{AOI in S.Sudan. Solid lines denote the annotated AOI while dashed lines denote the area covered by SSU-base and SSU-time pre-training data.}
    \label{map_SSU}
  \end{minipage}
\end{figure*}

% \begin{figure}[!t]
% \begin{center}
%     % \fbox{\rule{0pt}{2in} \rule{0.9\linewidth}{0pt}}
%   \includegraphics[width=\linewidth]{figures/geodata/germany.png}
% \end{center}
%   \caption{AOI in Ghana. Solid lines denote the AOI while washed lines denote the area covered by GHA-base pre-training data.}
% \label{DEU_data}
% \end{figure}

% \begin{figure}[!t]
% \begin{center}
%     % \fbox{\rule{0pt}{2in} \rule{0.9\linewidth}{0pt}}
%   \includegraphics[width=\linewidth]{figures/geodata/france.png}
% \end{center}
%   \caption{AOI in Ghana. Solid lines denote the AOI while washed lines denote the area covered by GHA-base pre-training data.}
% \label{FRA_data}
% \end{figure}

% \begin{figure}[!t]
% \begin{center}
%     % \fbox{\rule{0pt}{2in} \rule{0.9\linewidth}{0pt}}
%   \includegraphics[width=0.65\linewidth]{figures/geodata/ghana.png}
% \end{center}
%   \caption{AOI in Ghana. Solid lines denote the AOI while washed lines denote the area covered by GHA-base pre-training data.}
% \label{GHA_data}
% \end{figure}

% \begin{figure}[!t]
% \begin{center}
%     % \fbox{\rule{0pt}{2in} \rule{0.9\linewidth}{0pt}}
%   \includegraphics[width=1.0\linewidth]{figures/geodata/ssudan.png}
% \end{center}
%   \caption{AOI in S.Sudan. Solid lines denote the AOI while washed lines denote the area covered by GHA-base pre-training data.}
% \label{SSU_data}
% \end{figure}

Experiments are performed with publicly available data from four different AOI in Germany (DEU), France (FRA), Ghana (GHA) and S.Sudan (SSU). Annotated data were obtained from the original studies. For downloading and extracting data without annotations for self-supervised training we use the {\it DeepSatData} pipeline \cite{deepsatdata}. The choice of sample dimensions for the pre-training datasets (Table \ref{unsup_datasets}) was motivated by the sample size of the segmentation datasets (Table \ref{sup datasets}). It was desired that random crops during pre-training match the size of the respective segmentation datasets while maintaining balance between two desired properties: the need for enough ovelap between crops to obtain positive pairs and enough stochasticity when extracting different views. Samples for all datasets are square images.

In France we use the same training data as \cite{cscl} but keep only 13 out of the 166 available land cover types\footnote{https://www.data.gouv.fr/en/datasets/registre-parcellaire-graphique-rpg-contours-des-parcelles-et-ilots-culturaux-et-leur-groupe-de-cultures-majoritaire} (PPH, J6S, SNE, ORH, PRL, PTC, MH7, RDF, BOP, MLG, RVI, MIS, TRE) and only for year 2018. This choice was made on the basis of retaining the most numerous and best performing classes. This dataset is completely covered by {\it S2} tile T31TFM. The {\it FRA-base} dataset consists of {\it S2} tile T31TFM for year 2018, while the FRA-time dataset extends that for years 2016 and 2017. The FRA-space dataset consists of tiles T31TFN, T31TFM and T31TFL for year 2018.
In Germany we use the data from \cite{mtlcc} for year 2016 keeping all 17 crop type groupings (maize, wheat, meadow, winter barley, potato, rapeseed, summer barley, hop, triticale, oat, rye, sugar beet, spelt, aparagus, beans, peas, soybeans) from the  original study. The DEU-base dataset consists of tile T32UPU for year 2016 while the DEU-time dataset extends that with images from years 2017 and 2020. \\
Both datasets in Europe contain all 13 {\it Sentinel-2} bands plus an additional {\it day of year} (doy) parameter indicating the day of the year normalized in $(0, 1]$. Subsampling was performed uniformly in space such that even the smallest datasets cover the entire AOI. The locations of data for Germany and France are shown in Figs.\ref{map_DEU}, \ref{map_FRA}. % The locations of data in the evaluation set and subsampling of locations for the training sets can be seen in Figs. \ref{FRA_train_eval}, \ref{DEU_train_eval}.  

For experiments in Africa we used the Ghana and South Sudan datasets from \cite{Rustowicz2019SemanticSO} shared by the Radiant Earth Foundation\footnote{https://mlhub.earth/}. For Ghana we selected the 4 most popular crop types found in the AOI (ground nut, maize, rice and soy bean). The training and evaluation data span the 2016 growing season. The GHA-base dataset consists of tile T31PZR for year 2016. 
In South Sudan we selected 6 land cover types out of the most populated land cover types found in the data (bush, bare soil, forest, shrubland, grasslands and meadows, wetland) for the 2017 growing season. The SSU-base dataset consists of tile T35PMK for 2017 while the SSU-time dataset extends that to years 2018 and 2020. 
% The SSU-space dataset consists of tiles T35PMK, T35PKL, T35PKN for year 2017. 
Data in Ghana cover a relatively small AOI while data in South Sudan cover a very large AOI which spans the entire country. Both datasets are quite sparse both in terms of parcel locations (the average distance between parcels is large) but also in terms of ratio of foreground to total number of pixels (few foreground pixels w.r.t. total). The locations of data for Ghana and South Sudan are shown in Figs.\ref{map_GHA}, \ref{map_SSU}.
Both African datasets contain 10 {\it Sentinel-2} bands plus an additional doy parameter. We note that in order to protect privacy both datasets do not include information on the location of agricultural parcels. As such splitting the data into train and evaluation sets was performed randomly in space while ensuring that all datasets maintain an equal ratio from every class. We also note that for privacy preservation purposes the input satellite images in both datasets include some random noise component making them sub-optimal for image classification purposes.\\

%\begin{itemize}
%     \item  
% \end{itemize}
% From publicly available datasets we use Germany \cite{mtlcc} and France \cite{cscl}. Subsampling of the datasets occurs uniformly in space as seen in Fig.\ref{subsampling_data}. We always keep the evaluation set the same, as a result the eval set can be many times larger than the smaller training datasets, i.e. X sampels for Germany and X sampels for France.\\

% correct data sizes in parentheses

\begin{figure*}[!h]
  \centering
  \begin{minipage}[b]{0.475\textwidth}
    \centering
    \includegraphics[width=0.675\textwidth, trim={0 0 60 0}]{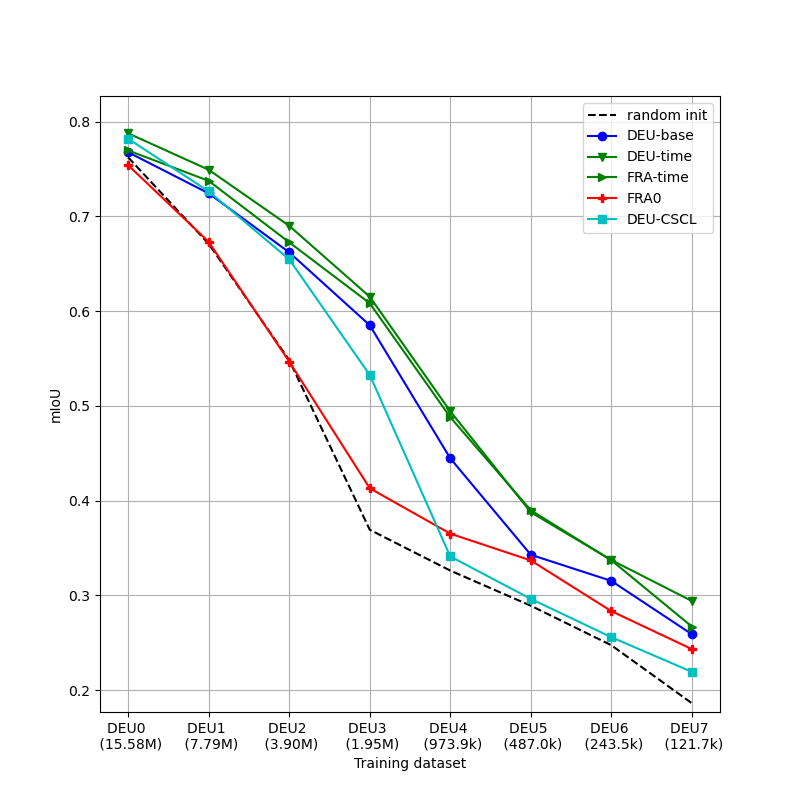}
    \caption{Germany - mIoU at evaluation set.}
    \label{DEU_scatter}
  \end{minipage}
  \hfill
  \begin{minipage}[b]{0.475\textwidth}
  % trim={<left> <lower> <right> <upper>}
    \centering
    \includegraphics[width=0.675\textwidth, trim={0 0 60 0}]{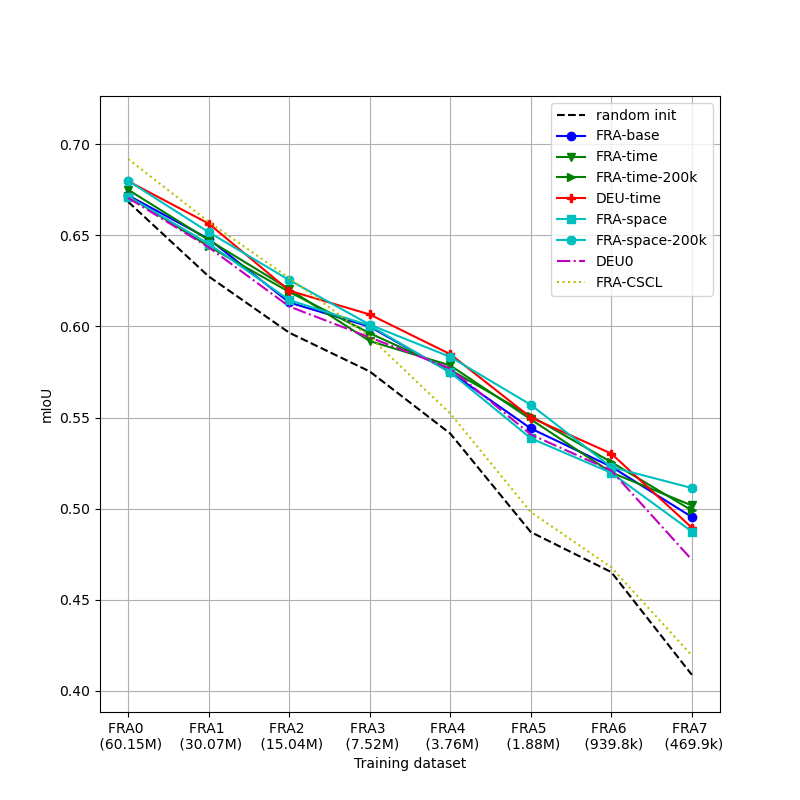}
    \caption{France - mIoU at evaluation set.}
    \label{FRA_scatter}
  \end{minipage}
\end{figure*}

\begin{figure*}[!h]
  \centering
  \begin{minipage}[b]{0.475\textwidth}
    \centering
    \includegraphics[width=0.675\textwidth, trim={0 0 60 0}]{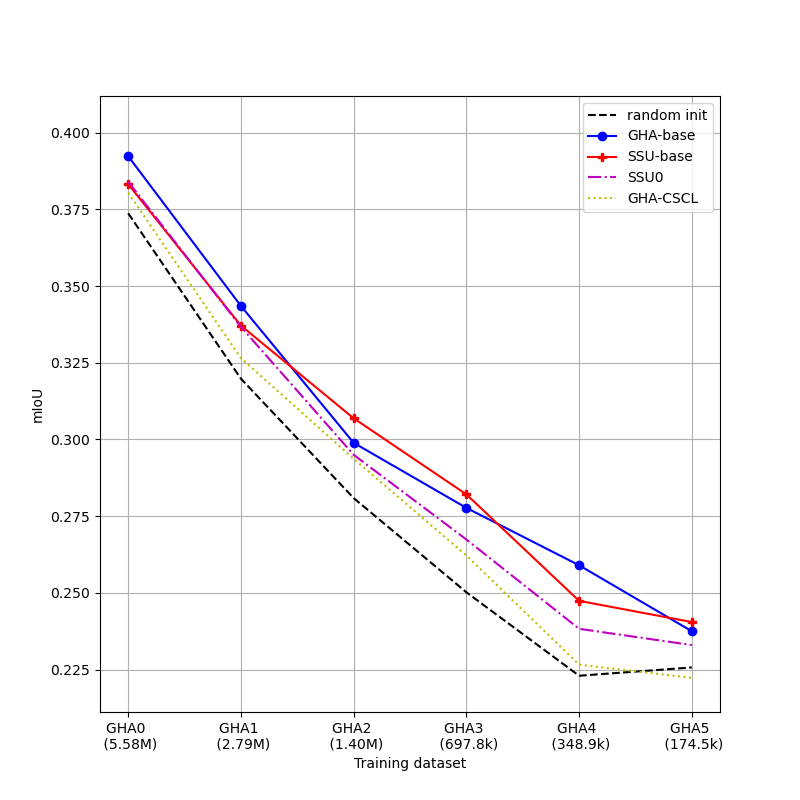}
    \caption{Ghana - mIoU at evaluation set.}
    \label{GHA_scatter}
  \end{minipage}
  \hfill
  \begin{minipage}[b]{0.475\textwidth}
  % trim={<left> <lower> <right> <upper>}
    \centering
    \includegraphics[width=0.675\textwidth, trim={0 0 60 0}]{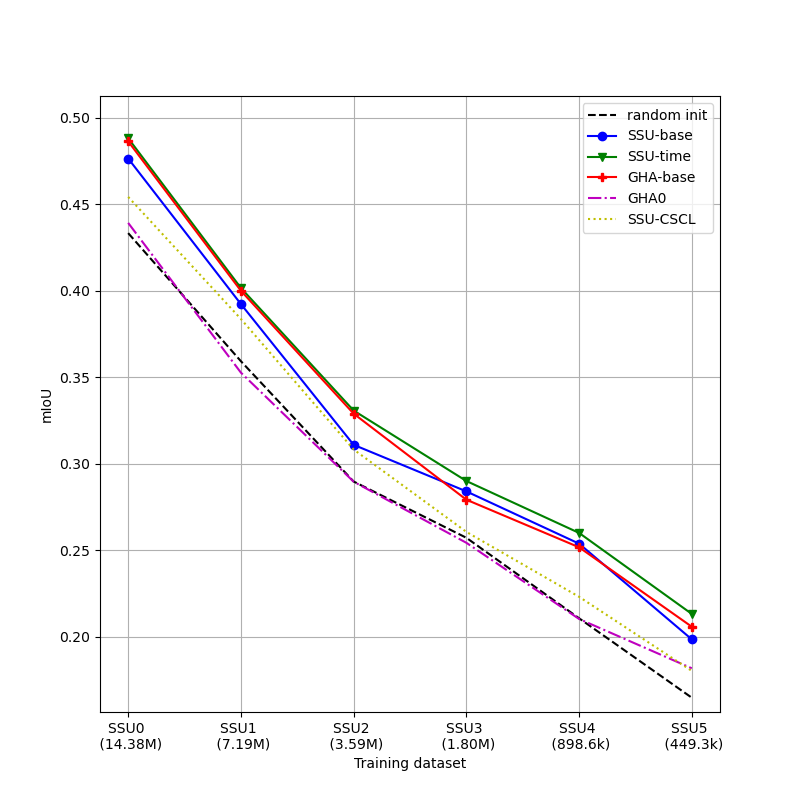}
    \caption{S.Sudan - mIoU at evaluation set.}
    \label{SSU_scatter}
  \end{minipage}
\end{figure*}

\subsection{Training implementation details}
For all experiments we used the {\it UNet3Df} models from \cite{cscl}. This model is a variation of {\it UNet3D} \cite{3dunet} - previously used for land cover semantic segmentation in \cite{Rustowicz2019SemanticSO, cscl} - better suited for metric style learning. We refer the reader to \cite{cscl} for further details. This model was preferred to {\it Conv-RNN} alternatives because it is faster to train thus, more suitable for use in multiple experiments. All models were trained using commodity Nvidia TITAN Xp and RTX 2080 gpus. A single card was used for training the final segmentation models. For pre-training two gpus were used in a data parallel fashion, however, the model can be pre-trained using a single gpu and this choice was employed to increase training speed rather than to overcome limitations imposed by accelerator memory. Further details can be found in Appendix \ref{training_details}.

\subsection{Comparison with random initialization}
We present a comparison between models starting with random parameter initialization and those pre-trained using the base datasets. Results are presented in Figs.\ref{DEU_scatter}, \ref{FRA_scatter}, \ref{GHA_scatter}, \ref{SSU_scatter} respectively for Germany, France, Ghana and S.Sudan. Overall, as expected, we observe evaluation performance to drop with decreasing number of samples in the training datasets. Additionally, pre-trained models are shown to outperform their counterparts in every case tested with no exception. From the horizontal distances between the pre-trained and random initialization curves it appears that pre-training has an equivalent effect with doubling the amount of training data. 
There is a general trend for pre-training to have a more significant benefit in the smaller annotation space. However, that is not always the case. For example in Germany performance seems to improve more on medium size (DEU3, DEU4) rather than the smallest datasets. 
We also note that pre-training performance gains are smaller on the African datasets. Potential reasons could be the inclusion of noise in the training datasets which introduces a gap between pre-training and training data or the large cloud cover which affects both pre-training and final segmentation training. 

\subsection{Comparison with state-of-the-art pre-training}
We compare EmbeddingEarth with the recently proposed CSCL method \cite{cscl} for supervised visual pre-training. While our method does not require ground truths for the pre-train task CSCL depends on provided annotations. As such, to ensure a fair comparison, CSCL pre-trained models only use the amount of annotations available in each training set, e.g. FRA2 segmentation model is initialized by a model pre-trained on FRA2. Results are presented in Figs.\ref{DEU_scatter}, \ref{FRA_scatter}, \ref{GHA_scatter}, \ref{SSU_scatter} respectively for Germany, France, Ghana and S.Sudan. We observe decreasing performance gains for CSCL in smaller size datasets which is expected given that the size of the pre-training set matches the training set size. Our method exhibits a much gentler performance degardation w.r.t. the dataset size and outperforms CSCL in every case apart from the largest datasets for France (FRA0-FRA2). 

% \begin{figure*}[!h]
%   \centering
%   \begin{minipage}[b]{0.475\textwidth}
%     \includegraphics[width=\textwidth, trim={0 0 0 0}]{figures/rand_vs_base/DEU_rand_vs_base.png}
%     \caption{Germany - mIoU at evaluation set. Comparison between random initialization vs DEU-base pre-training.}
%     \label{DEU_rand_vs_base}
%   \end{minipage}
%   \hfill
%   \begin{minipage}[b]{0.475\textwidth}
%   % trim={<left> <lower> <right> <upper>}
%     \includegraphics[width=\textwidth, trim={0 0 0 0}]{figures/rand_vs_base/FRA_rand_vs_base.png}
%     \caption{Germany - mIoU at evaluation set. Comparison between random initialization vs FRA-base pre-training.}
%     \label{FRA_rand_vs_base}
%   \end{minipage}
% \end{figure*}

% \begin{figure*}[!h]
%   \centering
%   \begin{minipage}[b]{0.475\textwidth}
%     \includegraphics[width=\textwidth, trim={0 0 0 0}]{figures/rand_vs_base/GHA_rand_vs_base.png}
%     \caption{Ghana - mIoU at evaluation set. Comparison between random initialization vs GHA-base pre-training.}
%     \label{GHA_rand_vs_base}
%   \end{minipage}
%   \hfill
%   \begin{minipage}[b]{0.475\textwidth}
%   % trim={<left> <lower> <right> <upper>}
%     \includegraphics[width=\textwidth, trim={0 0 0 0}]{figures/rand_vs_base/SSU_rand_vs_base.png}
%     \caption{S.Sudan - mIoU at evaluation set. Comparison between random initialization vs SSU-base pre-training.}
%     \label{SSU_rand_vs_base}
%   \end{minipage}
% \end{figure*}

\subsection{Analysis of results}
\begin{figure}[!h]
  \centering
  \begin{subfigure}{0.24\textwidth}
    \includegraphics[width=\textwidth, trim={20 2 60 0}]{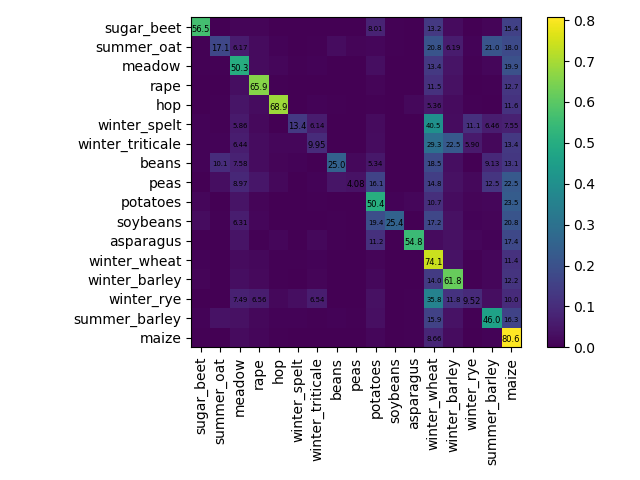}
    \caption{random init}
  \end{subfigure}
  \begin{subfigure}{0.24\textwidth}
    \includegraphics[width=\textwidth, trim={40 2 40 0}]{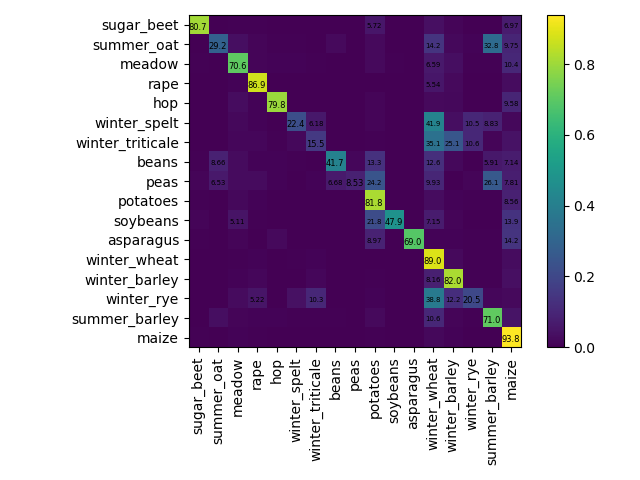}
    \caption{DEU-base pre-train}
  \end{subfigure}
  \caption{Germany - Confusion matrices at evaluation set for (a) random and (b) pre-trained initializations. Models trained using the DEU4 dataset.} % Better viewed in color and zoomed in.}
  \label{DEU4_conf}
\end{figure}

\begin{figure}[!t]
%  \begin{minipage}[b]{0.475\textwidth}
    \centering
    \includegraphics[width=0.475\textwidth, trim={0 0 0 0}]{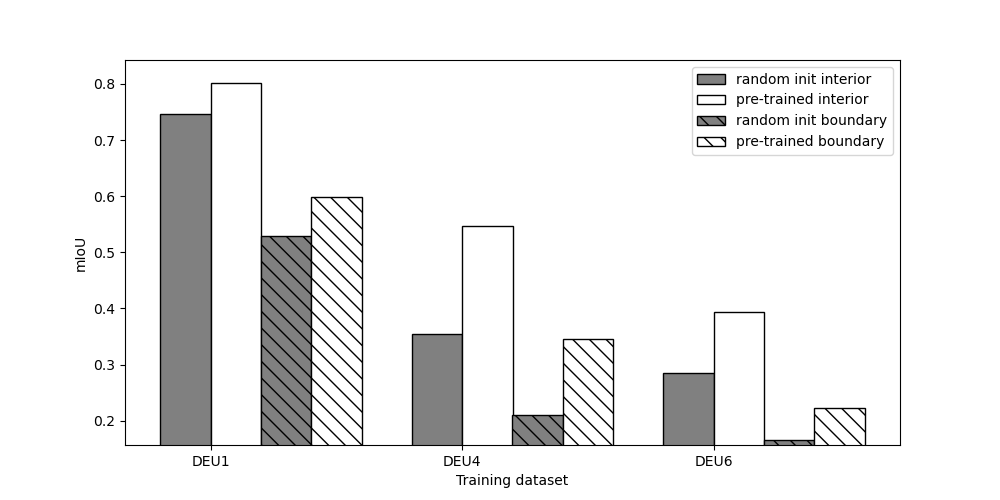}
    \caption{Germany - mIoU at evaluation set. Comparison between random initialization and pre-training for interior and boundary pixels.}
    \label{DEU_int_vs_bound0}
%   \end{minipage}
\end{figure}

We find that improvements are generally evenly distributed among classes and pixel location within a parcel. In Fig.\ref{DEU4_conf} we show confusion matrices for the DEU4 dataset and compare per class performance for random initialization of parameters and DEU-base pre-training. Similar confusion matrices for all countries can be found in Figs.\ref{DEU_conf}, \ref{FRA_conf}, \ref{GHA_conf}, \ref{SSU_conf} in Appendix \ref{appendix_eval_metrics}. We observe improvements for all classes in Germany and the majority of classes elsewhere. In France Ghana and S.Sudan there are few classes for which random initialization performs better, however, overall improvement is always higher for the pre-trained models.  We also distinguish between performance improvement at parcel interior vs boundary locations. For this purpose a boundary location is assumed to be any location which does not share the same class as all of its $3 \times 3$ local neighbourhood. Results are plotted in Fig.\ref{DEU_int_vs_bound0} for Germany. Additional plots for all countries can be found in Figs.\ref{DEU_int_vs_bound}, \ref{FRA_int_vs_bound}, \ref{GHA_int_vs_bound}, \ref{SSU_int_vs_bound} in Appendix \ref{appendix_eval_metrics}. We observe similar improvement at both interior and boundary locations which is different from the findings of \cite{cscl} whose pre-training task lead to greater performance gains at the boundaries. 

% explore where improvement comes from\\
% Present confusion matrices for the four datasets for random init and best pre-trained model for 2-3 datasets per country?.\\
% perform boundary-interior points analysis.\\

\subsection{Ablation studies}
In this section we present results from a series of ablation studies to asses how various qualities of the pre-training affect final segmentation performance. % and the potential applicability 
% \begin{figure}[!t]
% \begin{center}
%     % \fbox{\rule{0pt}{2in} \rule{0.9\linewidth}{0pt}}
%   \includegraphics[width=1.0\linewidth]{figures/ablations/FRA_temperature_ablation.png}
% \end{center}
%   \caption{Ablation on $\tau$ parameter (eq. \ref{losseq}) in France. All models were pretrained using the FRA-base dataset.}
% \label{ablation_temperature}
% \end{figure}
\begin{figure}[!t]
\begin{center}
  \includegraphics[width=0.7\linewidth]{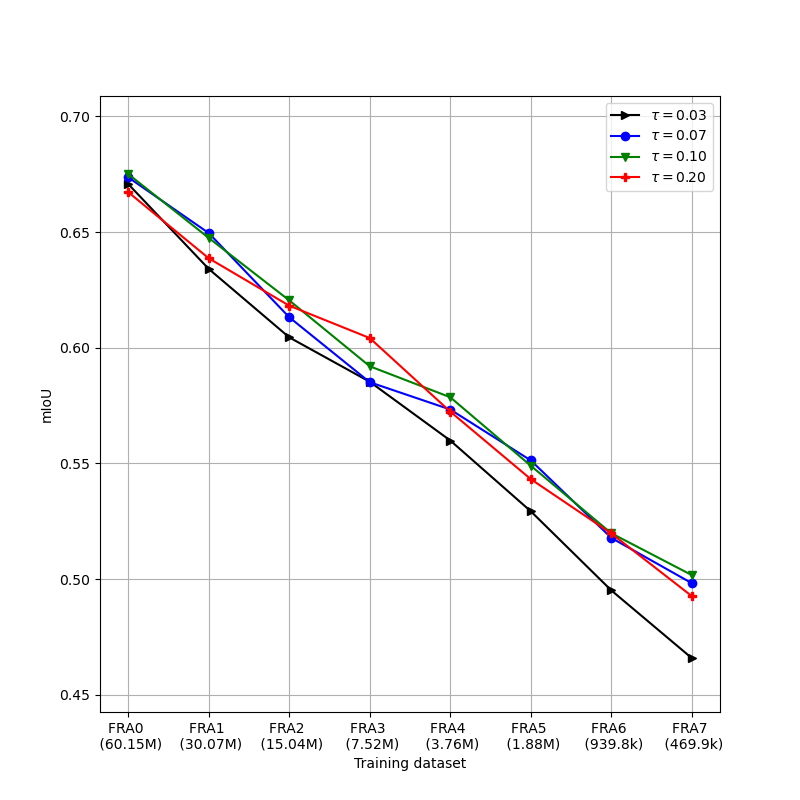}
\end{center}
  \caption{Ablation on $\tau$ parameter (eq. \ref{losseq}) in France. All models were pretrained using the FRA-base dataset.}
\label{ablation_temperature}
\end{figure}

For deciding on a value for the {\bf temperature scaling} parameter $\tau$ (eq. \ref{losseq}) we explore how various $\tau$ values affect performance. We test typically used values close to $0.07$ used in the original implementation of {\it MoCo} \cite{moco}. Results are shown in Fig.\ref{ablation_temperature}. Overall most temperature parameter values appear to lead to similar performance on semantic segmentation apart from $\tau=0.03$ which seems to significantly lag on smaller size datasets. The choice of $\tau=0.10$ for the remaining experiments was made because this parameter was found to be a good performer in most cases and is close to the values typically used in {\it MoCo}. 

\begin{figure}[!t]
\begin{center}
  \includegraphics[width=0.7\linewidth]{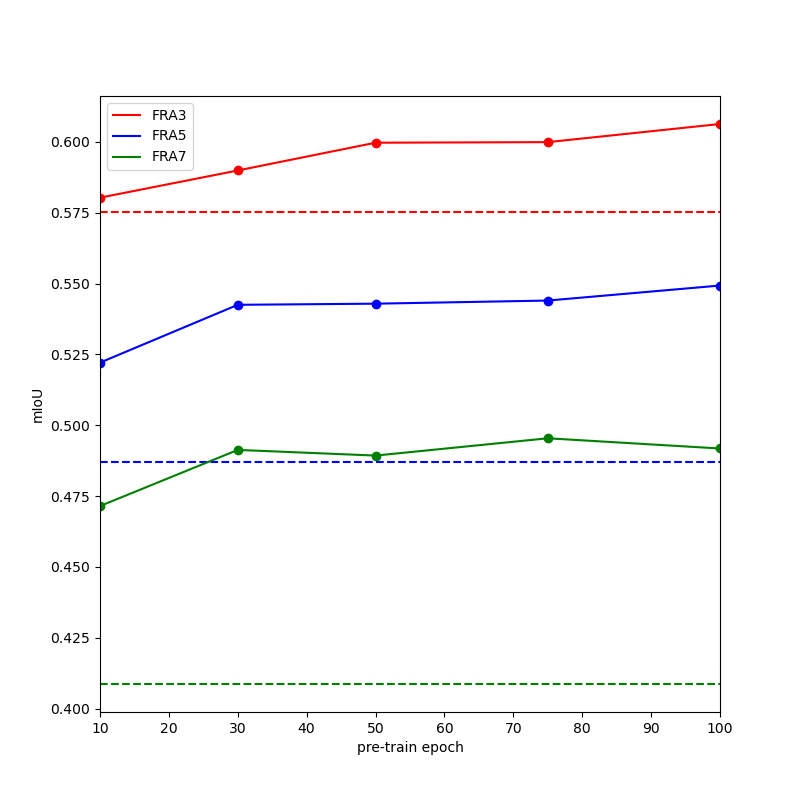}
\end{center}
  \caption{France - mIoU at evaluation set. Segmentation performance by sampling initialization points during pre-training. Dashed lines indicate random initialization baselines.}
\label{pretrain_epochs}
\end{figure}

It is of interest to find what is the {\bf optimum number of pre-training steps}. To find out we are sampling initialization values at various stages during pre-training and compare downstream segmentation performance. Results are presented in Fig.\ref{pretrain_epochs}. Overall we observe that even minimal pre-training for $10$ epochs improves over random initialization in every case and performance continues to improve with more pre-training. This suggests that the selected number of $100$ epochs might be sub-optimal and more steps could lead to better performance in practice. 

% \begin{figure*}[!h]
%   \centering
%   \begin{minipage}[b]{0.475\textwidth}
%     \includegraphics[width=\textwidth, trim={0 0 0 0}]{figures/ablations/DEU_DEU_FRA_pretrain.png}
%     \caption{Germany - mIoU at evaluation set. Comparison between DEU-time and FRA-time pre-training.}
%     \label{DEU_DEU_vs_FRA}
%   \end{minipage}
%   \hfill
%   \begin{minipage}[b]{0.475\textwidth}
%   % trim={<left> <lower> <right> <upper>}
%     \includegraphics[width=\textwidth, trim={0 0 0 0}]{figures/ablations/FRA_DEU_FRA_pretrain.png}
%     \caption{France - mIoU at evaluation set. Comparison between FRA-time and DEU-time pre-training.}
%     \label{FRA_FRA_vs_DEU}
%   \end{minipage}
% \end{figure*}

% \begin{figure*}[!h]
%   \centering
%   \begin{minipage}[b]{0.475\textwidth}
%     \includegraphics[width=\textwidth, trim={0 0 0 0}]{figures/ablations/GHA_GHA_SSU_pretrain.png}
%     \caption{Ghana - mIoU at evaluation set. Comparison between GHA-base and SSU-base pre-training.}
%     \label{GHA_GHA_vs_SSU}
%   \end{minipage}
%   \hfill
%   \begin{minipage}[b]{0.475\textwidth}
%   % trim={<left> <lower> <right> <upper>}
%     \includegraphics[width=\textwidth, trim={0 0 0 0}]{figures/ablations/SSU_SSU_GHA_pretrain.png}
%     \caption{S.Sudan - mIoU at evaluation set. Comparison between SSU-base and GHA-base pre-training.}
%     \label{SSU_SSU_vs_GHA}
%   \end{minipage}
% \end{figure*}

Next we want to assess the {\bf generality of pretrained features}. By mixing pre-training and finetuning regions we assess the importance of pre-training at a different location than that of the downstream task. More specifically we swap the initializations between each set of countries in Europe and Africa.
Results using the swaped datasets for Germany, France, Ghana and S.Sudan can be seen in Figs.\ref{DEU_scatter}, \ref{FRA_scatter} and Figs.\ref{GHA_scatter}, \ref{SSU_scatter}.
% and Africa and varrying amounts of training annotations can be seen in Tables \ref{DEU_results_FRApretrain}, \ref{FRA_results_DEUpretrain}, \ref{GHA_results_SSUpretrain}, \ref{SSU_results_GHApretrain} respectively for Germany, France, Ghana and S.Sudan.
We find that swapping locations has little to no effect over downstream segmentation performance and in numerous cases we obtain improved performance using an initialization point from a different country. 
This suggests that the reason our pre-training scheme improves performance is not just because the network learns the instance discrimination task for a particular region. This is also in contrast with the tendency of fully supervised land cover segmentation models not to generalize to completely new regions. To our knowledge this is the first method exhibiting that quality. More importantly it opens the possibility to pre-train and collect a set of initialization values for popular land cover identification models and use these instead of random initialization for downstream tasks. This method has been very successfully applied to other computer vision problems such as object detection and semantic segmentation using ImageNet \cite{imagenet} pre-trained initializations for faster training and improved performance. 

% Variations in pretrain and train image size appear not to affect performance.\\
% \begin{figure*}[!h]
%   \centering
%   \begin{minipage}[b]{0.475\textwidth}
%     \includegraphics[width=\textwidth, trim={0 0 0 0}]{figures/ablations/DEU_base_time.png}
%     \caption{Germany - mIoU at evaluation set. Comparison between DEU-base and DEU-time pre-training.}
%     \label{DEU_base_vs_time}
%   \end{minipage}
%   \hfill
%   \begin{minipage}[b]{0.475\textwidth}
%   % trim={<left> <lower> <right> <upper>}
%     \includegraphics[width=\textwidth, trim={0 0 0 0}]{IEEEtran/figures/ablations/FRA_base_space_time.png}
%     \caption{France - mIoU at evaluation set. Comparison between FRA-base, FRA-space and FRA-time pre-training.}
%     \label{FRA_base_vs_space_vs_time}
%   \end{minipage}
% \end{figure*}

\begin{figure}[!h]
\begin{center}
  \includegraphics[width=0.7\linewidth]{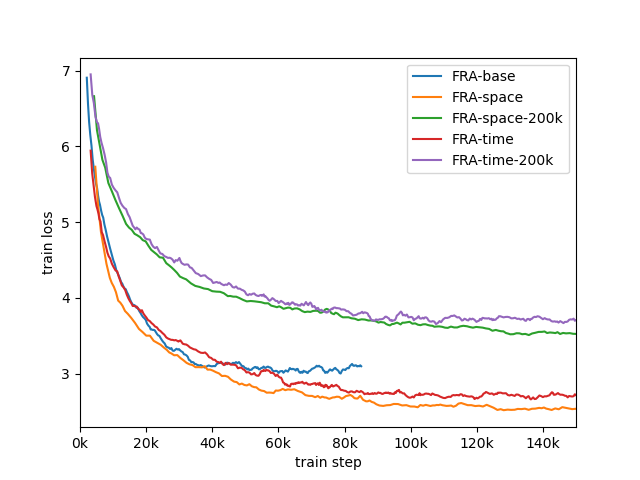}
\end{center}
  \caption{France. Train loss during pre-training.}
\label{pretrain_loss_FRA}
\end{figure}

So far we have seen that we get stronger improvements compared to the randomly initialized baselines for small size training datasets. In a similar spirit, here we assess the significance of the {\bf size of pre-training datasets} in segmentation performance. We compare the results obtained using the {\it base} datasets with the {\it time} and {\it space} datasets from Table \ref{unsup_datasets}. Additionally, we want to asses the relative importance of using more locations for a given period of time compared to a longer period for a given set of locations. Results are presented in Figs.\ref{DEU_scatter}, \ref{FRA_scatter}. We note a general trend for {\it time} pre-trained models to outperform both {\it base} and {\it space} pre-trained ones when using the default 67k data queue. Using a larger pre-training AOI does not appear to offer any benefits and models pre-trained with FRA-space are in most cases outperformed by FRA-base models. This indicates that pre-training over a larger AOI than that of the training set (Fig.\ref{map_FRA}) can hinder downstream model performance. However, it was previously shown that pre-training at a different location does not affect performance which raises the question why using a pre-train dataset covering larger regions does not offer any improvement. One reason could be that the size of the examined datasets is already large enough to achieve most of the benefits of pre-training. 
Another possibility is that the larger the area covered by a pre-training dataset the easier the objective function becomes to optimize as queue encodings correspond to increasingly disparate locations and are easier to distinguish from the query vectors. We present some empirical evidence towards the latter reason in Fig.\ref{pretrain_loss_FRA} where we plot the train loss for pre-training in France and observe lower values for the FRA-space dataset consistently during training indicating an easier optimization objective. Furthermore, we perform pre-training by tripling the size of the data queue to size 201k. Increasing the size of the queue introduces a tradeoff between including more diversity in the negative samples and increasing the duration (in training steps) an encoding spends in the queue.
More diversity is achieved simply because encodings from more locations are included in the queue and thus as negative pairs in the loss function (eq.\ref{losseq}). The number of negative pairs has been previously shown to improve the quality of learnt features is contrastive learning \cite{moco, simclr}. However, increasing the duration an encoding spends in the data queue means that currently encoded queries are being compared with encodings from earlier instantiations of the keys encoder which might not be as relevant during the current training step. In our case training with a 201k data queue is found to be a more difficult pre-training optimization objective (Fig.\ref{pretrain_loss_FRA}). From Fig.\ref{FRA_scatter} we observe  an improvement in performance for the {\it space} dataset but not for the {\it time} dataset.    

% \begin{figure}[!h]
% \begin{center}
%     % \fbox{\rule{0pt}{2in} \rule{0.9\linewidth}{0pt}}
%   \includegraphics[width=1.0\linewidth]{figures/ablations/DEU_unsup_vs_sup_pretrain.png}
% \end{center}
%   \caption{Germany - mIoU at evaluation set. Comparison between DEU-time, DEU-base and FRA0 supervised pre-training.}
% \label{sup_vs_unsup_pretrain}
% \end{figure}
We have shown the benefits of the proposed pre-training scheme w.r.t. random initialization. Another option is to use a previously trained {\bf supervised land cover segmentation model as an initialization point} for a new region. To recreate a realistic scenario in which we have a large labelled dataset in a source country and we want to train for a target country we use the best model trained with the largest dataset in each respective country, i.e. DEU0, FRA0, GHA0, SSU0, as the initialization for another country in the same continent. Results are shown in Figs.\ref{DEU_scatter}, \ref{FRA_scatter}, \ref{GHA_scatter}, \ref{SSU_scatter}. In all cases self-supervised initializations outperform the fully supervised ones. In terms of the observed differences the results are mixed. For Germany and S.Sudan the differences with self-supervision are significant, the fully supervised initializations are closer to the random rather than the self-supervised ones. For France and Ghana the differences are not as significant. 

% in Germany for three different initialization schemes: random, using the best model trained on the FRA0 dataset and models pre-trained with self-supervision in Germany or France. 

In most cases, we observe a smaller benefit of supervised pre-training over random initialization for the larger training datasets and some benefit for the smaller datasets. An exception to pattern is Ghana in which the benefit is almost constant for all dataset sizes. %In every case supervised pre-training is outperformed by our method.  

\section{Conclusion}
In this paper we presented a self-supervised pre-training methodology for land cover semantic segmentation tasks using time series of satellite images. To assess the merits of our approach we performed a large scale study using {\it Sentinel-2} images from four countries in Europe and Africa. In every case tested we found significant improvements up to $+25\%$ absolute {\it mIoU} over random initialization and showed improvements over fully supervised pre-training. Additionally we have shown that the learnt features generalize equally well to unseen regions significantly improving downstream segmentation performance in different countries than the ones used during pre-training.
Because of this we see our work opening a new direction in land cover segmentation making it possible to use a set of pre-trained model weights for high performing segmentation models as initialization points for EO related downstream tasks similar to how pre-trained models have been used in computer vision tasks. The {\it Embedding Earth} pretext task does not use any annotations, thus, it is not in any way restricted to learning a particular type or land cover but is a system for extracting features characteristic of the spatiotemporal patterns of the input image time series. As a result, although not tested in tasks other than land cover segmentation, we expect pre-trained initialization points to be beneficial for other EO tasks including but not limited to dense classification such as object detection and dense regression. We leave these possibilities as a topic for future research.

\appendices
\begin{figure*}[!ht]
  \centering
  \begin{minipage}[b]{0.475\textwidth}
  % trim={<left> <lower> <right> <upper>}
    \includegraphics[trim={0 27 0 0}, width=\textwidth]{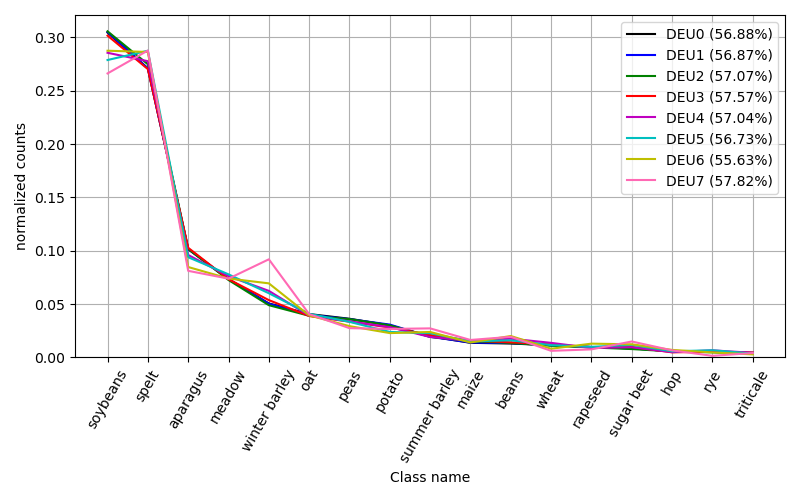}
    % \caption{AOI in France. Solid lines denote the AOI while dashed blue lines denote the area covered by FRA-space pre-training data. Note that FRA-base and FRA-time datasets cover the same area.}
    % \label{class_DEU}
  \end{minipage}
  \begin{minipage}[b]{0.475\textwidth}
    \centering
    \includegraphics[trim={0 27 0 0}, width=\textwidth]{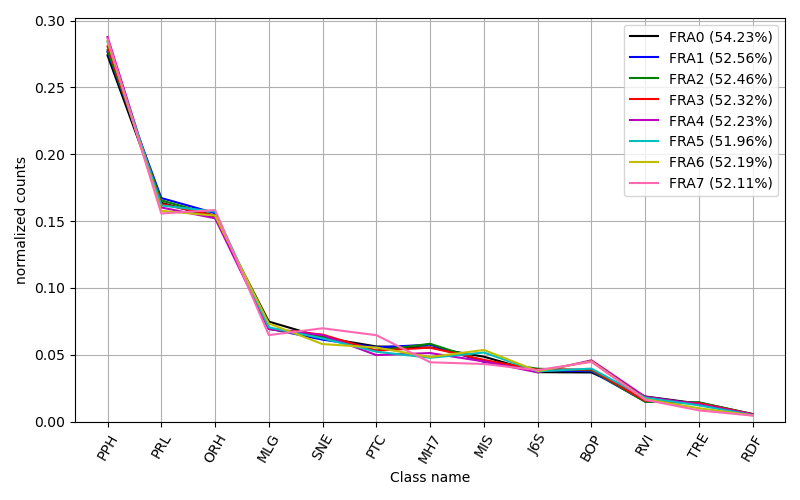}
    % \label{map_DEU}
  \end{minipage}
  \centering
  \begin{minipage}[b]{0.475\textwidth}
  % trim={<left> <lower> <right> <upper>}
    \includegraphics[trim={0 30 0 0}, width=\textwidth]{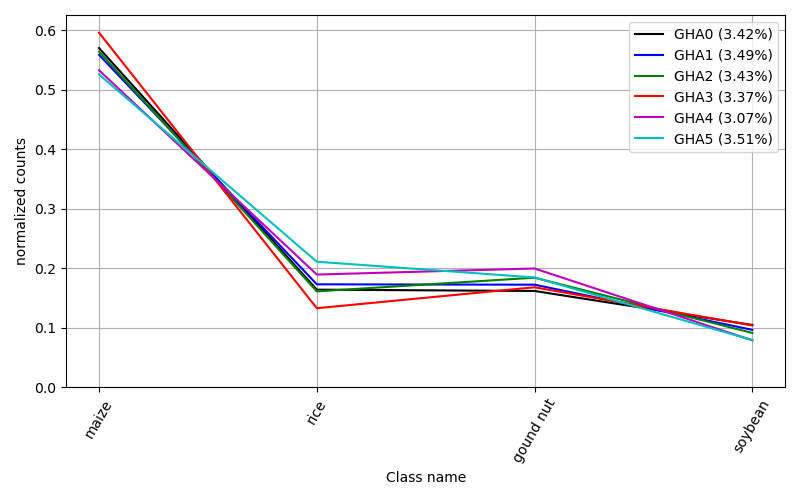}
    % \caption{AOI in France. Solid lines denote the AOI while dashed blue lines denote the area covered by FRA-space pre-training data. Note that FRA-base and FRA-time datasets cover the same area.}
    % \label{map_FRA}
  \end{minipage}
  \hfill
  \begin{minipage}[b]{0.475\textwidth}
  % trim={<left> <lower> <right> <upper>}
    \includegraphics[trim={0 30 0 0}, width=\textwidth]{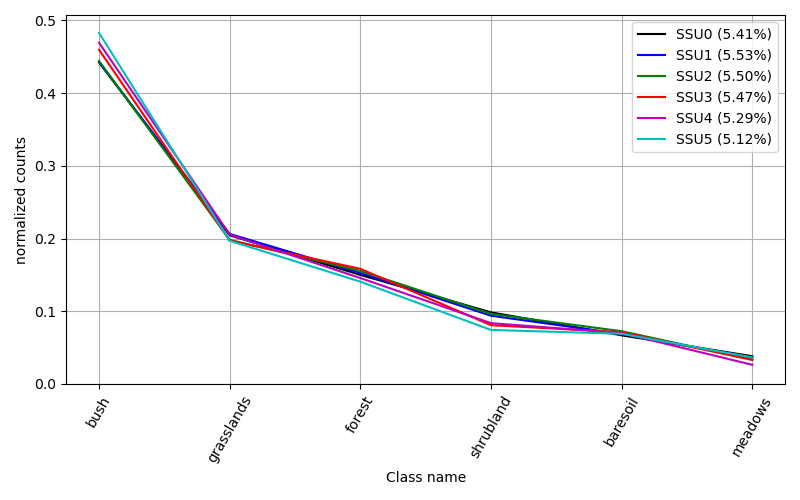}
    % \caption{AOI in France. Solid lines denote the AOI while dashed blue lines denote the area covered by FRA-space pre-training data. Note that FRA-base and FRA-time datasets cover the same area.}
    % \label{map_FRA}
  \end{minipage}
  \caption{Class distribution among the training datasets. (top-left) Germany, (top-right) France, (bottom-left) Ghana, (bottom-right) S.Sudan. $\%$ in legend parentheses is the ratio of foreground to overall pixels for each dataset.}
  \label{class_distr}
\end{figure*}

\section{Dataset class distributions}
In Fig.\ref{class_distr} we present normalized pixel counts (pixel count for each class divided by total) for all training datasets in the four countries tested. By design we maintain a similar distribution of pixel counts is maintained for all datasets in each country. Only smaller size datasets exhibit some limited variation from the average distribution which can be attributed to their limited size, i.e. winter barley in DEU7. In the legend we also show the ratio of foreground to overall pixels for each dataset. Note how this value is significantly lower for the Arican compared to the European datasets as a result of the data sparsity in Ghana and S.SUdan.

\section{Training implementation details} \label{training_details}
{\bf Pre-training details.} All models were pre-trained using the SGD optimizer with momentum $0.9$ and a learning rate $0.0001$ reduced by $\times 10$ at epochs $40$ and $60$. Unless otherwise indicated the temperature parameter for eq.\ref{losseq} is set to $0.10$, the size of the keys queue is set to $67,200$ and we always sampled $32$ pixels in correspondence between views to use as positive pairs.
The batch size was set to $60$ for datasets in Europe and $30$ for datasets in Africa. All models were pre-trained for $100$ epochs in Europe and $200$ epochs in Africa. We chose to train for more epochs in Africa because the number of samples in the pre-training set was reduced as a consequence of the larger sample size ($100$ pixels compared to $48$ for France and $32$ for Germany). 
%During pre-training we used random crops of sizes $24$, $32$ for Germany and France and $64$ for Africa in an effort to best match . 
The crop size during pre-training was selected such that it matched the sample size of the respective segmentation dataset. France is an exception to this rule as models there were pre-trained with crop size $32$ while the dataset from \cite{cscl} contains samples of size $48$. We have empirically found this not to be significant for final segmentation performance as there are cases where models pre-trained in Germany with much smaller crop size outperform models pre-trained in France.     
{\bf Segmentation training details.} All models were trained using the Adam optimizer \cite{adam} with default hyperparameters $\beta_1=0.9, \beta_2=0.999$ and a learning rate $0.0001$ reduced exponentialy at a rate of $0.975$ every two epochs. Details on the total number of epochs and batch sizes used can be found in Table \ref{train_epochs_batch}. In all cases training hyperparameters were kept constant between runs using the same dataset, e.g. all models trained with the FRA1 dataset use the same training protocol. 

\begin{table}[!ht]
\begin{center}
\begin{tabular}{|c|c|c|}
\hline
train set & N epochs & batch size \\
\hline \hline
DEU0-3 & 150 & 32 \\
DEU4-5 & 225 & 16 \\
DEU6-7 & 225 & 8 \\
\hline \hline
FRA0-3 & 150 & 32 \\
FRA4-5 & 225 & 16 \\
FRA6-7 & 225 & 8 \\
\hline \hline
GHA0 & 400 & 16 \\
GHA1-5 & 400 & 8 \\
\hline \hline
SSU0-1 & 300 & 16 \\
SSU2-5 & 400 & 8 \\
\hline
\end{tabular}
\end{center}
\caption{Segmentation training. Number of epochs and batch size.}
\label{train_epochs_batch}
\end{table}

\section{Detailed evaluation metrics} \label{appendix_eval_metrics}
\begin{figure}[!h]
  \begin{subfigure}{0.24\textwidth}
    \includegraphics[width=\textwidth, trim={20 2 10 0}]{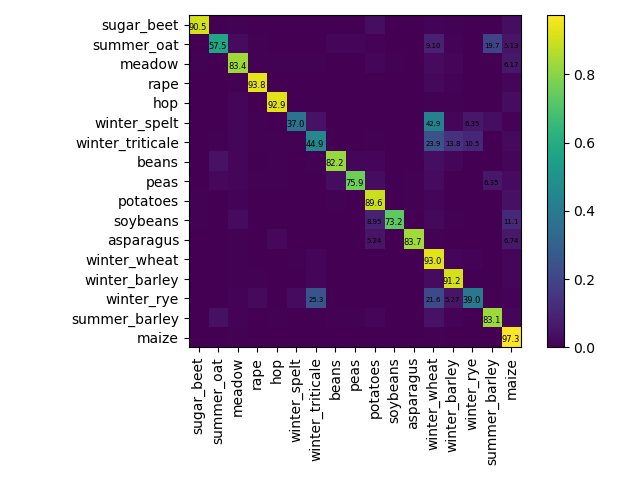}
    \caption{DEU1-random}
  \end{subfigure}
  \begin{subfigure}{0.24\textwidth}
    \includegraphics[width=\textwidth, trim={20 2 10 0}]{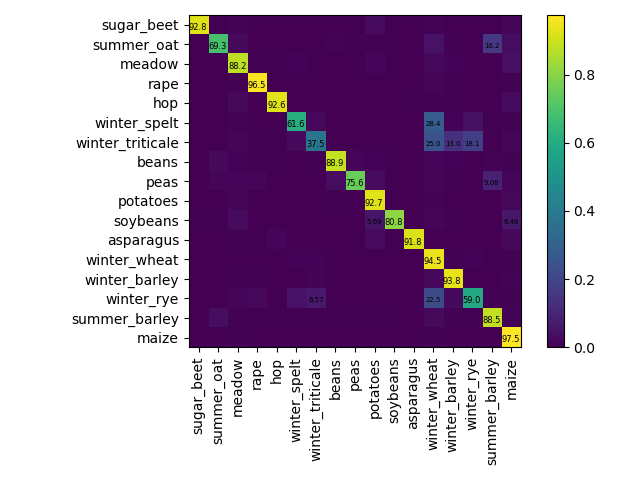}
    \caption{DEU1-pretrain}
  \end{subfigure}
  \begin{subfigure}{0.24\textwidth}
    \includegraphics[width=\textwidth, trim={20 2 10 0}]{figures/confmat/Germany/base_2e4_all.png}
    \caption{DEU4-random}
  \end{subfigure}
  \begin{subfigure}{0.24\textwidth}
    \includegraphics[width=\textwidth, trim={20 2 10 0}]{figures/confmat/Germany/moco_2e4_all.png}
    \caption{DEU4-pretrain}
  \end{subfigure}
  \begin{subfigure}{0.24\textwidth}
    \includegraphics[width=\textwidth, trim={20 2 10 0}]{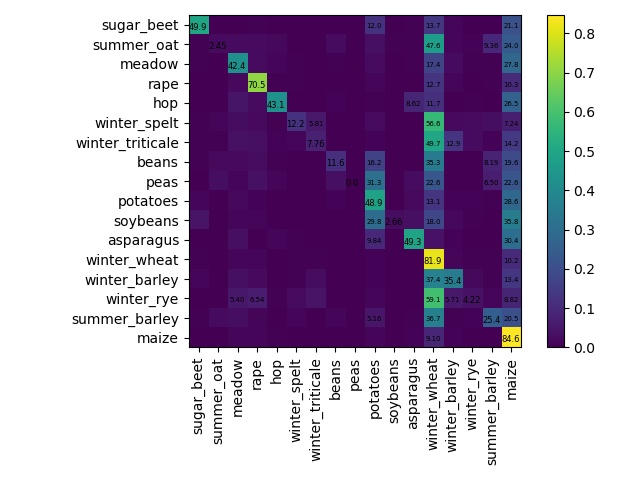}
    \caption{DEU6-random}
  \end{subfigure}
  \begin{subfigure}{0.24\textwidth}
    \includegraphics[width=\textwidth, trim={20 2 10 0}]{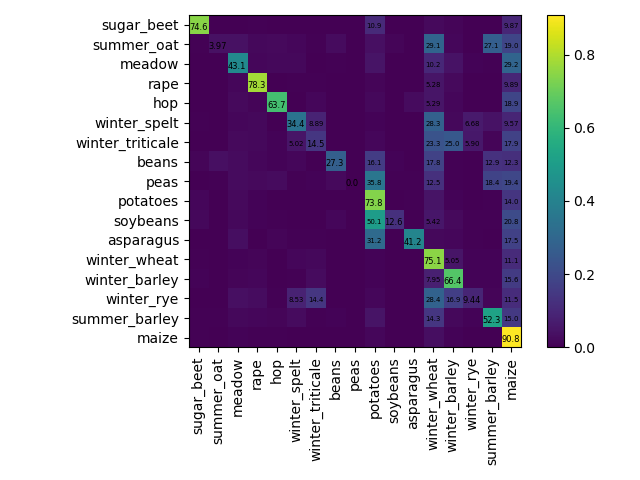}
    \caption{DEU6-pretrain}
  \end{subfigure}
  \caption{Germany - Confusion matrices at evaluation set. From top to bottom: DEU1, DEU4, DEU6 training datasets, left to right: random and pre-trained initialization. Better viewed in color and zoomed in.}
  \label{DEU_conf}
\end{figure}

\begin{figure}[!h]
  \begin{subfigure}{0.24\textwidth}
    \includegraphics[width=\textwidth, trim={0 0 0 0}]{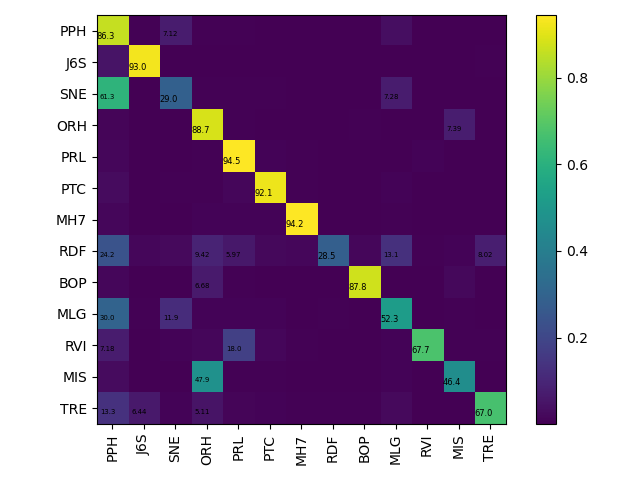}
    \caption{FRA1-random}
  \end{subfigure}
  \begin{subfigure}{0.24\textwidth}
    \includegraphics[width=\textwidth, trim={0 0 0 0}]{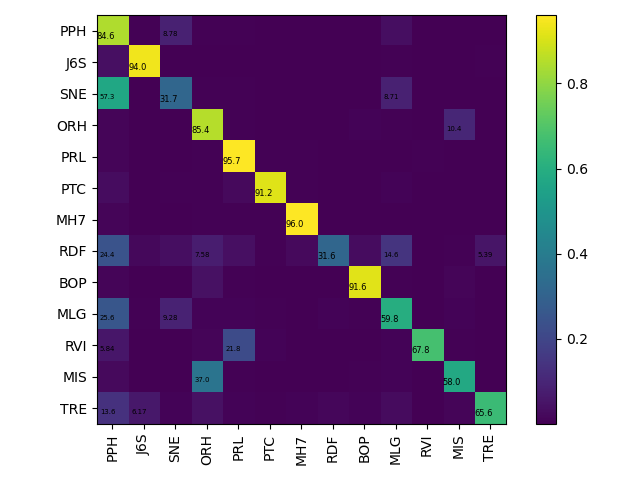}
    \caption{FRA1-pretrain}
  \end{subfigure}
  \begin{subfigure}{0.24\textwidth}
    \includegraphics[width=\textwidth, trim={0 0 0 0}]{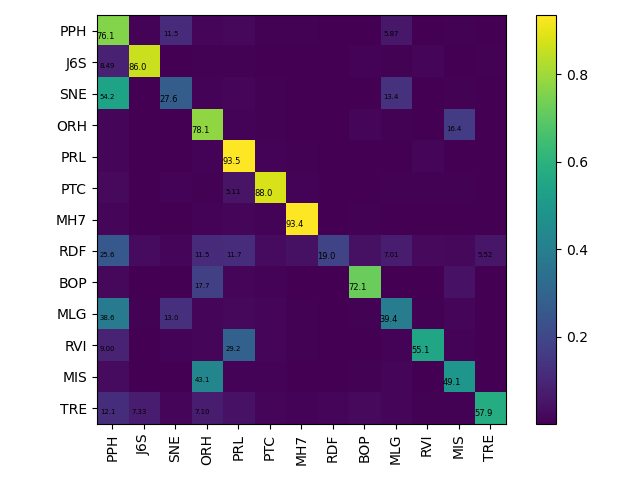}
    \caption{FRA4-random}
  \end{subfigure}
  \begin{subfigure}{0.24\textwidth}
    \includegraphics[width=\textwidth, trim={0 0 0 0}]{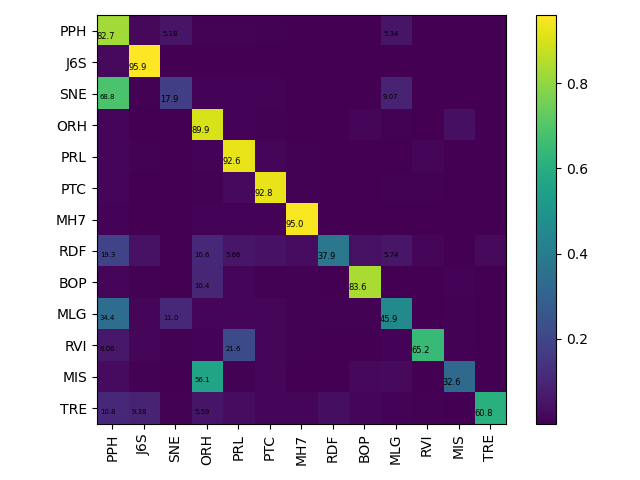}
    \caption{FRA4-pretrain}
  \end{subfigure}
  \begin{subfigure}{0.24\textwidth}
    \includegraphics[width=\textwidth, trim={0 0 0 0}]{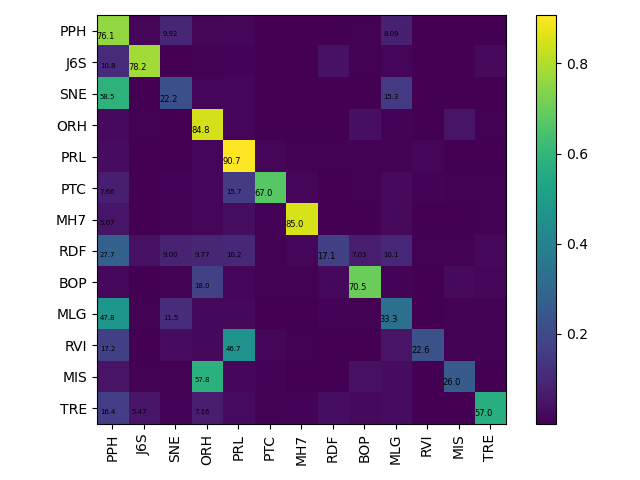}
    \caption{FRA6-random}
  \end{subfigure}
  \begin{subfigure}{0.24\textwidth}
    \includegraphics[width=\textwidth, trim={0 0 0 0}]{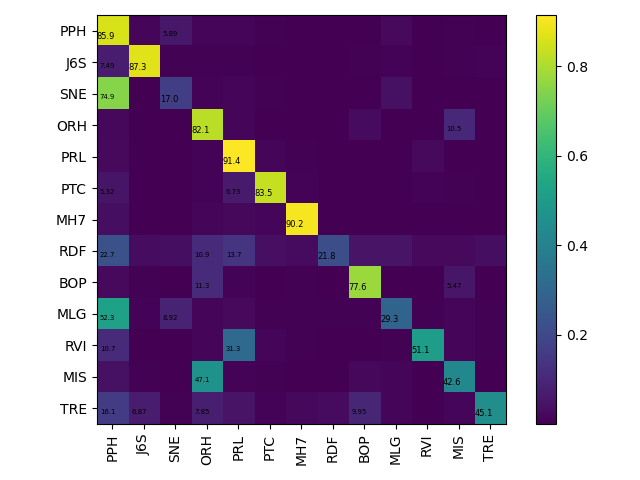}
    \caption{FRA6-pretrain}
  \end{subfigure}
  \caption{France - Confusion matrices at evaluation set. From top to bottom: FRA1, FRA4, FRA6 training datasets, left to right: random and pre-trained initialization. Better viewed in color and zoomed in.}
  \label{FRA_conf}
\end{figure}

\begin{figure}[!h]
  \begin{subfigure}{0.24\textwidth}
    \includegraphics[width=0.75\textwidth, trim={0 0 0
    0}]{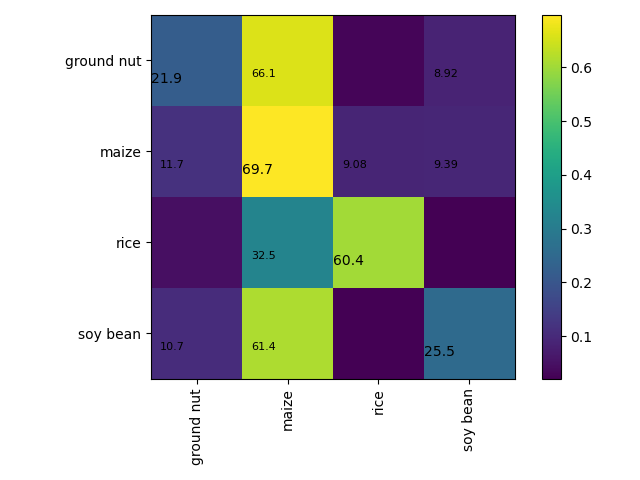}
    \caption{GHA1-random}
  \end{subfigure}
  \begin{subfigure}{0.24\textwidth}
    \includegraphics[width=0.75\textwidth, trim={0 0 0
    0}]{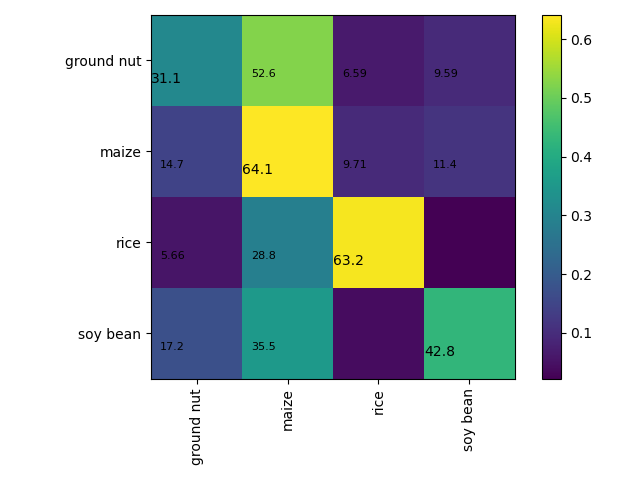}
    \caption{GHA1-pretrain}
  \end{subfigure}
  \begin{subfigure}{0.24\textwidth}
    \includegraphics[width=0.75\textwidth, trim={20 2 10 0}]{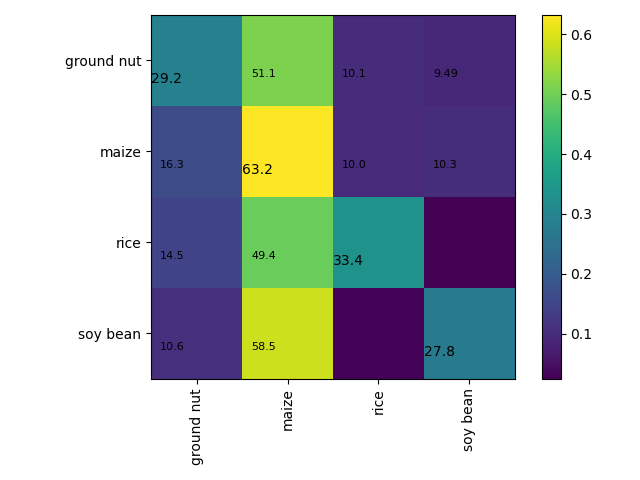}
    \caption{GHA3-random}
  \end{subfigure}
  \begin{subfigure}{0.24\textwidth}
    \includegraphics[width=0.75\textwidth, trim={20 2 10 0}]{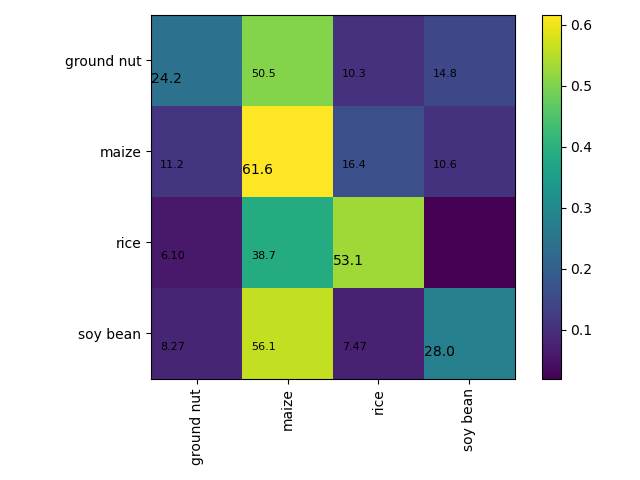}
    \caption{GHA3-pretrain}
  \end{subfigure}
  \begin{subfigure}{0.24\textwidth}
    \includegraphics[width=0.75\textwidth, trim={20 2 10 0}]{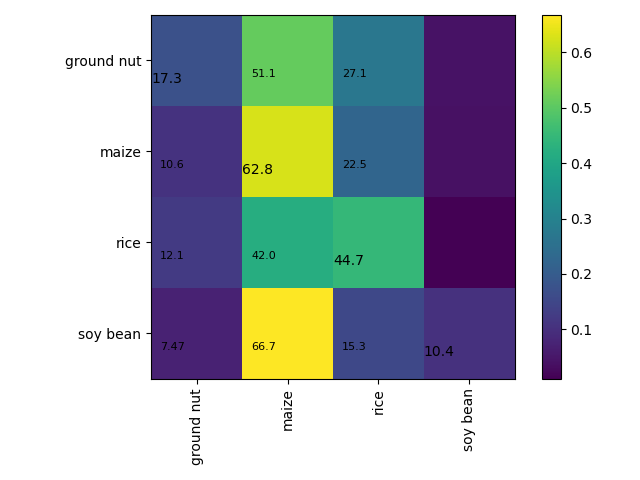}
    \caption{GHA5-random}
  \end{subfigure}
  \begin{subfigure}{0.24\textwidth}
    \includegraphics[width=0.75\textwidth, trim={20 2 10 0}]{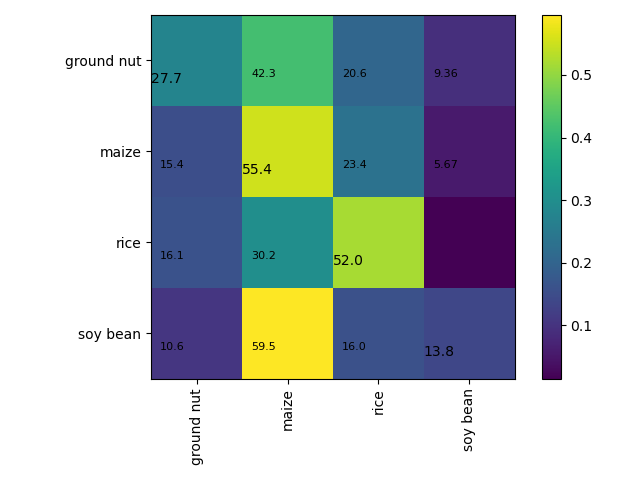}
    \caption{GHA5-pretrain}
  \end{subfigure}
  \caption{Ghana - Confusion matrices at evaluation set. From top to bottom: GHA1, GHA3, GHA5 training datasets, left to right: random and pre-trained initialization. Better viewed in color and zoomed in.}
  \label{GHA_conf}
\end{figure}

\begin{figure}[!h]
  \begin{subfigure}{0.24\textwidth}
    \includegraphics[width=\textwidth, trim={0 0 0
    0}]{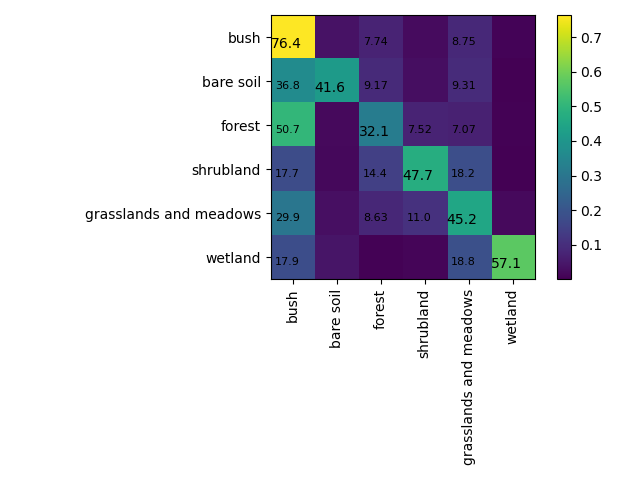}
    \caption{SSU1-random}
  \end{subfigure}
  \begin{subfigure}{0.24\textwidth}
    \includegraphics[width=\textwidth, trim={0 0 0
    0}]{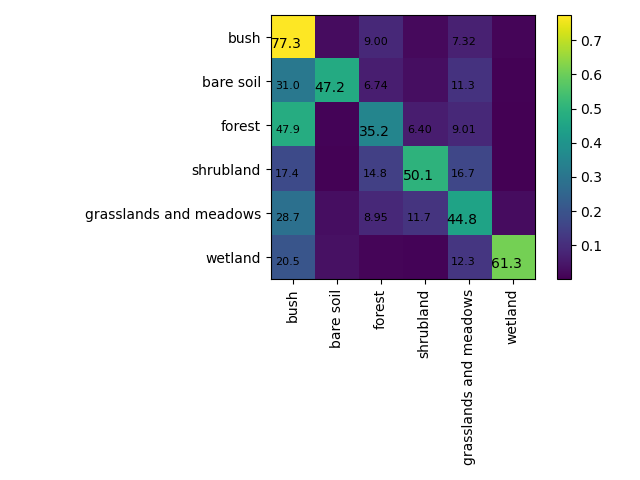}
    \caption{SSU1-pretrain}
  \end{subfigure}
  \begin{subfigure}{0.24\textwidth}
    \includegraphics[width=\textwidth, trim={20 2 10 0}]{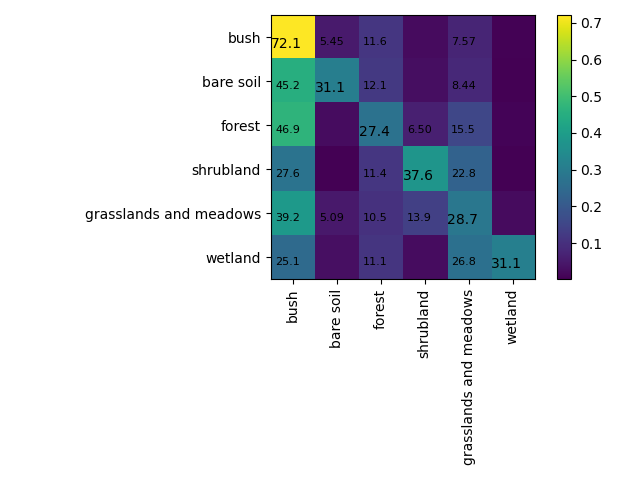}
    \caption{SSU3-random}
  \end{subfigure}
  \begin{subfigure}{0.24\textwidth}
    \includegraphics[width=\textwidth, trim={20 2 10 0}]{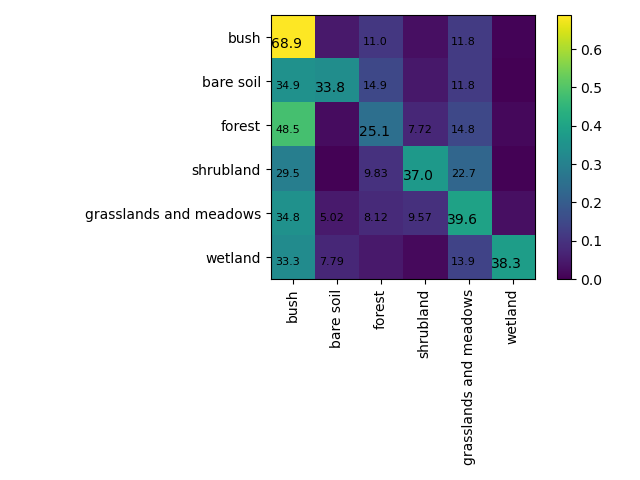}
    \caption{SSU3-pretrain}
  \end{subfigure}
  \begin{subfigure}{0.24\textwidth}
    \includegraphics[width=\textwidth, trim={20 2 10 0}]{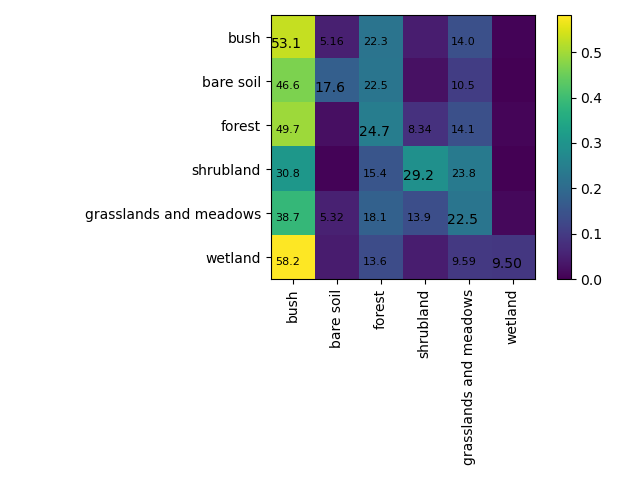}
    \caption{SSU5-random}
  \end{subfigure}
  \begin{subfigure}{0.24\textwidth}
    \includegraphics[width=\textwidth, trim={20 2 10 0}]{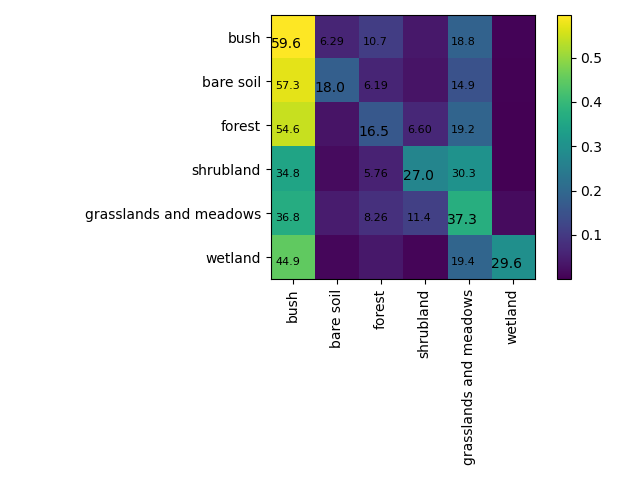}
    \caption{SSU5-pretrain}
  \end{subfigure}
  \caption{S.Sudan - Confusion matrices at evaluation set. From top to bottom: SSU1, SSU3, SSU5 training datasets, left to right: random and pre-trained initialization. Better viewed in color and zoomed in.}
  \label{SSU_conf}
\end{figure}

\begin{figure}[!h]
  \centering
%   \begin{minipage}[b]{0.475\textwidth}
    \includegraphics[width=0.475\textwidth, trim={0 0 0 0}]{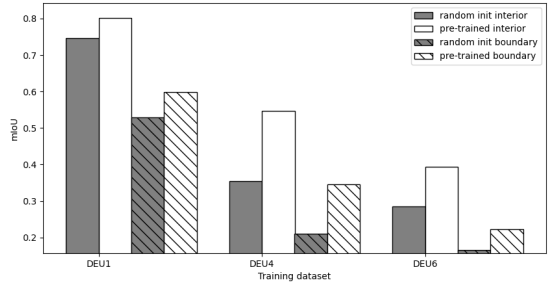}
    \caption{Germany - mIoU at evaluation set. Comparison between random initialization and pre-training for interior and boundary pixels.}
    \label{DEU_int_vs_bound}
%   \end{minipage}
%   \hfill
\end{figure}

\begin{figure}[!h]
%   \begin{minipage}[b]{0.475\textwidth}
  % trim={<left> <lower> <right> <upper>}
    \includegraphics[width=0.475\textwidth, trim={0 0 0 0}]{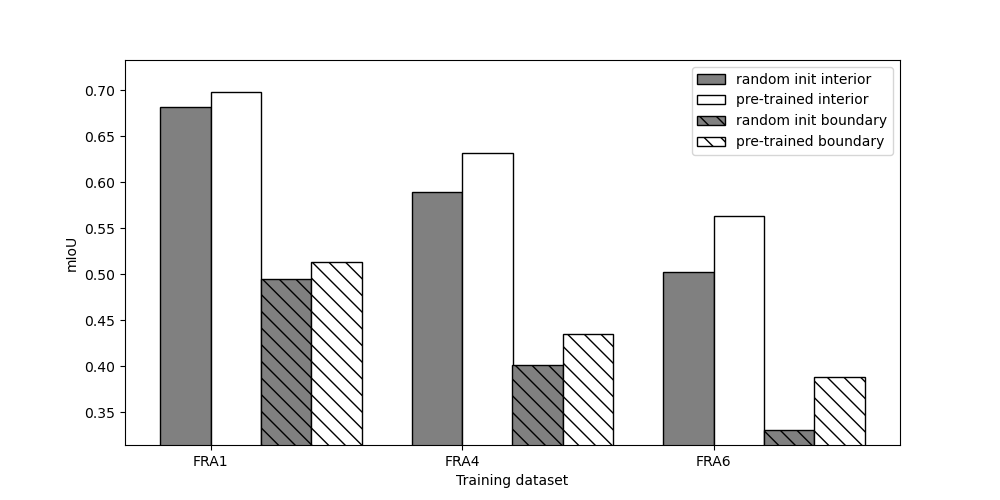}
    \caption{France - mIoU at evaluation set. Comparison between random initialization and pre-training for interior and boundary pixels.}
    \label{FRA_int_vs_bound}
%   \end{minipage}
\end{figure}

\begin{figure}[!h]
  \centering
%   \begin{minipage}[b]{0.475\textwidth}
    \includegraphics[width=0.475\textwidth, trim={0 0 0 0}]{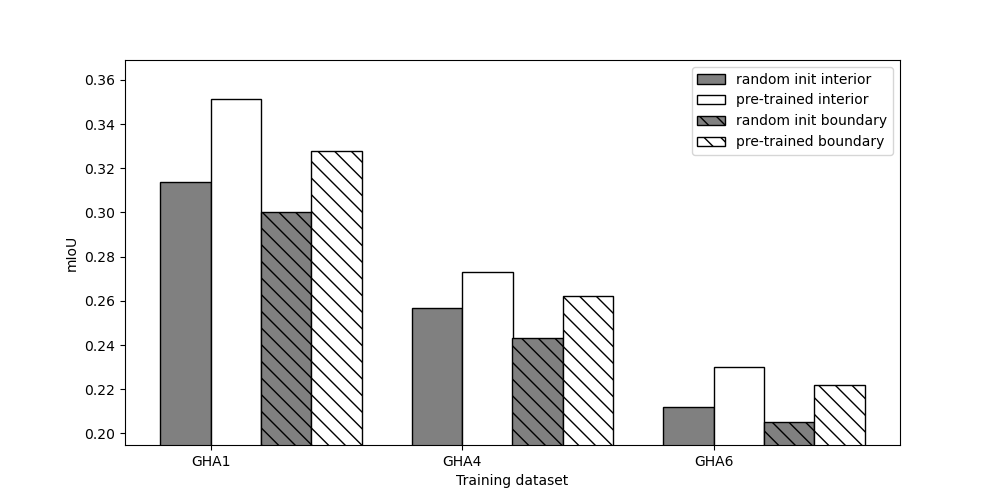}
    \caption{Ghana - mIoU at evaluation set. Comparison between random initialization and pre-training for interior and boundary pixels.}
    \label{GHA_int_vs_bound}
%   \end{minipage}
%   \hfill
%   \begin{minipage}[b]{0.475\textwidth}
  % trim={<left> <lower> <right> <upper>}
\end{figure}

\begin{figure}[!h]
    \includegraphics[width=0.475\textwidth, trim={0 0 0 0}]{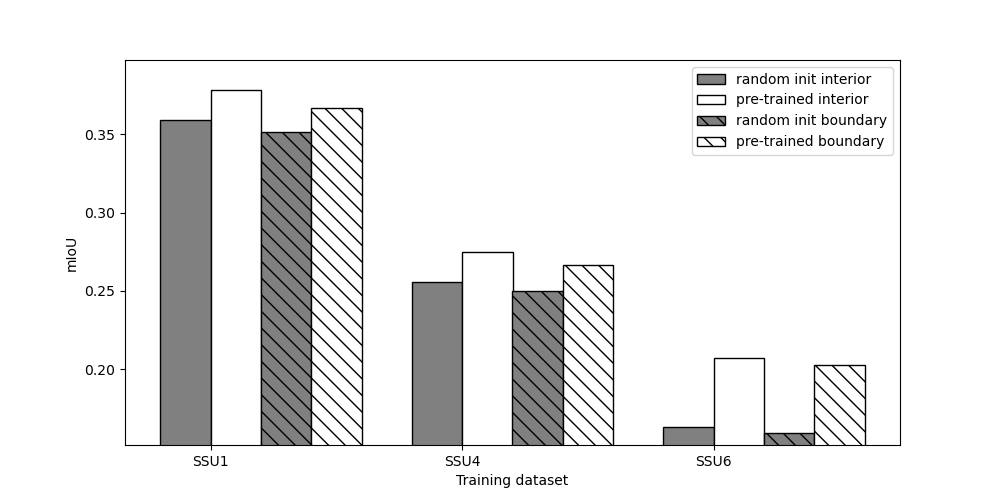}
    \caption{S.Sudan - mIoU at evaluation set. Comparison between random initialization and pre-training for interior and boundary pixels.}
    \label{SSU_int_vs_bound}
%   \end{minipage}
\end{figure}

Tables \ref{DEU_results}, \ref{GHA_results}, \ref{FRA_results}, \ref{SSU_results} show detailed evaluation metrics for all runs in Germany, Ghana, France and S.Sudan. We present macro (averaged over classes) {\it accuracy}, {\it mIoU} and {\it F1} scores and micro (averaged over pixels) {\it accuracy} and {\it mIoU}. Figs.\ref{DEU_conf}, \ref{FRA_conf}, \ref{GHA_conf}, \ref{SSU_conf} show confusion matrices for Germany, France, Ghana and S.Sudan respectively. All values of the diagonal are displayed. To avoid cluttering non diagonal values are only displayed if they are greater than $0.05$. Figs. \ref{DEU_int_vs_bound}, \ref{FRA_int_vs_bound}, \ref{GHA_int_vs_bound} \ref{SSU_int_vs_bound} present a performance comparison between random and pre-trained initialization for object interior and boundary pixels for Germany, France, Ghana and S.SUdan. 

\begin{table}[!h]
\begin{center}
\begin{tabular}{|c|c|c|c|c|c|c|c|c|}
\hline
\multicolumn{2}{|c|}{datasets} & \multicolumn{3}{|c|}{macro} & \multicolumn{2}{|c|}{micro}\\
\hline
pre-train & train & acc & iou & F1 & acc & iou\\
\hline \hline
- &	DEU0 & 0.8377 & 0.7623 & 0.8524 & 0.9261 & 0.8624\\
DEU-base & DEU0 & 0.8435 & 0.7677 & 0.8566 & 0.9233 & 0.868\\
DEU-time & DEU0 & 0.8625 & 0.7877 & 0.8717 & 0.9341 & 0.8763\\
FRA-time & DEU0 & 0.8537 & 0.7696 & 0.8597 & 0.9291 & 0.8676\\
$FRA0^{*}$ & DEU0 & 0.8289 & 0.7540 & 0.8463 & 0.9218 & 0.8549 \\
DEU0$^{\dagger}$ & DEU0 & 0.8588 & 0.7819 & 0.8667 & 0.9334 & 0.8755\\
\hline
- &	DEU1 & 0.7535 & 0.6710 & 0.7810 & 0.8956 &	0.8105 \\
DEU-base & DEU1 & 0.8058 & 0.7243 & 0.8225 & 0.9160 & 0.8449\\
DEU-time & DEU1 & 0.8286 & 0.7492 &	0.8422 & 0.9228 & 0.8567 \\
FRA-time & DEU1 & 0.8184 & 0.7375 & 0.8349 & 0.9176	& 0.8477 \\
FRA0 & DEU1 & 0.7719 & 0.6731 & 0.7867 & 0.8925 & 0.8058 \\
DEU1$^{\dagger}$ & DEU1 & 0.811 & 0.7264 & 0.8249 & 0.9173 & 0.8472\\
\hline
- &	DEU2 & 0.6740 & 0.5482 & 0.6819 & 0.8341 &	0.7154 \\
DEU-base & DEU2 & 0.7476 & 0.6620 & 0.7745 & 0.8952 & 0.8103\\
DEU-time & DEU2 & 0.7848 & 0.6900 & 0.7948 & 0.9043 & 0.8254 \\
FRA-time & DEU2 & 0.7616 & 0.6726 & 0.7816 & 0.8978	& 0.8146 \\
FRA0 & DEU2 & 0.6546 & 0.5467 & 0.6851 & 0.8278 & 0.7062 \\
DEU2$^{\dagger}$ & DEU2 & 0.751 & 0.6546 & 0.767 & 0.8935 & 0.8075\\
\hline
- &	DEU3 & 0.4904 & 0.3692 & 0.5137 &	0.6977 & 0.5358 \\
DEU-base & DEU3 & 0.6910 & 0.5851 & 0.7100 & 0.8631 & 0.7592\\
DEU-time & DEU3 &  0.7050 & 0.6152 & 0.7320 & 0.8766 & 0.7804 \\
FRA-time & DEU3 & 0.7028 & 0.6083 & 0.7312 & 0.8691	& 0.7685 \\
FRA0 & DEU3 & 0.5305 & 0.4130 & 0.5609 & 0.7309 & 0.5796 \\
DEU3$^{\dagger}$ & DEU3 & 0.6416 & 0.5326 & 0.6621 & 0.8422 & 0.7274\\
\hline
- &	DEU4 & 0.4292 & 0.3261 & 0.4611 & 0.6730 & 0.5072 \\
DEU-base & DEU4 & 0.5440 & 0.4446 & 0.5733 & 0.7578 & 0.6516\\
DEU-time & DEU4 & 0.5920 & 0.4943 & 0.6198 & 0.8206 & 0.6957 \\
FRA-time & DEU4 & 0.5904 & 0.4880 & 0.6201 & 0.8056 & 0.6745 \\
FRA0 & DEU4 & 0.4673 & 0.3651 & 0.5016 & 0.7132 & 0.5542 \\
DEU4$^{\dagger}$ & DEU4 & 0.4372 & 0.341 & 0.4729 & 0.7068 & 0.5466\\
\hline
- &	DEU5 & 0.4042 & 0.289 & 0.4150 & 0.6377 & 0.4682 \\
DEU-base & DEU5 & 0.4403 & 0.3428 & 0.4774 & 0.6969 & 0.5358\\
DEU-time & DEU5 & 0.4961 & 0.3877 &	0.5217 & 0.7350 & 0.5711 \\
FRA-time & DEU5 & 0.4849 & 0.3898 & 0.5243 & 0.7269	& 0.5710 \\
FRA0 & DEU5 & 0.4208 & 0.3368 & 0.4678 & 0.6991 & 0.5374 \\
DEU5$^{\dagger}$ & DEU5 & 0.3889 & 0.2962 & 0.4481 & 0.6546 & 0.4866\\
\hline
- &	DEU6 & 0.3369 & 0.2473 & 0.3853 & 0.6254 & 0.4528 \\
DEU-base & DEU6 & 0.4171 & 0.3152 & 0.4419 & 0.6781 & 0.5129\\
DEU-time & DEU6 & 0.4441 & 0.3372 &	0.4888 & 0.7057 & 0.5453 \\
FRA-time & DEU6 & 0.4315 & 0.3370 & 0.4732 & 0.7108	& 0.5514 \\
FRA0 & DEU6 & 0.364 & 0.2833 & 0.4250 & 0.6727 & 0.5068 \\
DEU6$^{\dagger}$ & DEU6 & 0.3493 & 0.2558 & 0.3899 & 0.6513 & 0.4829\\
\hline
- &	DEU7 & 0.2818 & 0.1861 & 0.3361 & 0.5779 & 0.4063 \\
DEU-base & DEU7 & 0.3514 & 0.2589 & 0.3694 & 0.6553 & 0.4873\\
DEU-time & DEU7 & 0.3820 & 0.2940 & 0.4230 & 0.6540 & 0.4854 \\
FRA-time & DEU7 & 0.3517 & 0.2670 & 0.4012 & 0.6502 & 0.4817 \\
FRA0 & DEU7 & 0.3331 & 0.2432 & 0.3687 & 0.6471 & 0.4783 \\
DEU7$^{\dagger}$ & DEU7 & 0.3007 & 0.2192 & 0.3831 & 0.6228 & 0.4522\\
\hline
\end{tabular}
\end{center}
\caption{Results in Germany. * supervised pre-training, $\dagger$ CSCL pre-training \cite{cscl}}
\label{DEU_results}
\end{table}

\begin{table}[!h]
\begin{center}
\begin{tabular}{|c|c|c|c|c|c|c|c|c|}
\hline
\multicolumn{2}{|c|}{datasets} & \multicolumn{3}{|c|}{macro} & \multicolumn{2}{|c|}{micro}\\
\hline
pre-train & train & acc & iou & F1 & acc & iou\\
\hline \hline
- &	FRA0 & 0.7592 & 0.6685 & 0.7733 & 0.8333 & 0.7142\\
FRA-base & FRA0 & 0.7647 & 0.6720 & 0.7766 & 0.8345 & 0.7158\\
FRA-time & FRA0 &  0.7633 & 0.6751 & 0.7769 & 0.8352 & 0.7230\\
FRA-time* & FRA0 & 0.7668 & 0.6716 & 0.7777 & 0.8307 & 0.7104\\
FRA-space & FRA0 & 0.7652 & 0.6714 & 0.7763 & 0.8344 & 0.7159\\
FRA-space* & FRA0 & 0.7620 & 0.6800 & 0.7848 & 0.8373 & 72.01 \\
DEU-time & FRA0 &  0.7646 & 0.6798 & 0.7810 & 0.8349 & 0.7241\\
DEU0 & FRA0 & 0.7540 & 0.6706 & 0.7719 & 0.8312 & 0.7236 \\
FRA0$^{\dagger}$ & FRA0 & 0.773 & 0.692 & 0.7913 & 0.8421 & 0.7273\\

\hline
- &	FRA1 & 0.7292 & 0.6275 & 0.7394 & 0.8069 & 0.6773\\
FRA-base & FRA1 & 0.7498 & 0.6481 & 0.7585 & 0.8158 & 0.6889\\
FRA-time & FRA1 &  0.7452 & 0.6475 & 0.7552 & 0.823 & 0.6992\\
FRA-time* & FRA1 & 0.7482 & 0.644 & 0.756 & 0.7116 & 0.683\\
FRA-space & FRA1 & 0.7327 & 0.6454 & 0.7561 & 0.8172 & 0.6909 \\
FRA-space* & FRA1 & 0.754 & 0.6518 & 0.7619 & 0.8173 & 0.691 \\
DEU-time & FRA1 &  0.7487 & 0.6565 & 0.7639 & 0.8265 & 0.7043\\
DEU0 & FRA1 & 0.7373 & 0.6435 & 0.7514 & 0.8106 & 0.6837 \\
FRA1$^{\dagger}$ & FRA1 & 0.7515 & 0.6575 & 0.7611 & 0.8221 & 0.698\\

\hline
- &	FRA2 & 0.7008 & 0.5966 & 0.7089 & 0.7826 & 0.6469 \\
FRA-base & FRA2 & 0.6878 & 0.581 & 0.6999 & 0.7811 & 0.6408\\
FRA-time & FRA2 & 0.7294 & 0.6205 & 0.7363 & 0.7997 & 0.6630\\
FRA-time* & FRA2 &  0.7301 & 0.619 & 0.7289 & 0.7858 & 0.6502\\
FRA-space & FRA2 & 0.718 & 0.6145 & 0.7303 & 0.7930 & 0.6570\\
FRA-space* & FRA2 & 0.7338 & 0.6255 & 0.739 & 0.7941 & 0.6582\\
DEU-time & FRA2 &  0.7212 & 0.6197 & 0.7349 & 0.7977 & 0.6619\\
DEU0 & FRA2 & 0.7127 & 0.6110 & 0.7232 & 0.7840 & 0.6483 \\
FRA2$^{\dagger}$ & FRA2 & 0.7229 & 0.6266 & 0.7369 & 0.8066 & 0.6759\\

\hline
- &	FRA3 & 0.6713 & 0.5753 & 0.6941 & 0.7760 & 0.6339 \\
FRA-base & FRA3 & 0.6931 & 0.5999 & 0.7167 & 0.7948 & 0.6595\\
FRA-time & FRA3 & 0.6873 & 0.5920 & 0.7092 & 0.7899 & 0.6528\\
FRA-time* & FRA3 & 0.703 & 0.5963 & 0.7159 & 0.7758 & 0.6337\\
FRA-space & FRA3 & 0.7000 & 0.6005 & 0.7201 & 0.7852 & 0.6463 \\
FRA-space* & FRA3 & 0.7004 & 0.601 & 0.718 & 0.7888 & 0.6499\\
DEU-time & FRA3 &  0.7093 & 0.6066 & 0.7224 & 0.7823 & 0.6425\\
DEU0 & FRA3 & 0.6951 & 0.5940 & 0.7104 & 0.7750 & 0.6343 \\
FRA3$^{\dagger}$ & FRA3 & 0.6862 & 0.5943 & 0.7091 & 0.7877 & 0.6498\\

\hline
- &	FRA4 & 0.6426 & 0.5412 & 0.6632 & 0.7560 & 0.6077\\
FRA-base & FRA4 & 0.6767 & 0.5749 & 0.6973 & 0.7760 & 0.6340\\
FRA-time & FRA4 & 0.6835 & 0.5786 & 0.6992 & 0.7796 & 0.6388\\
FRA-time* & FRA4 & 0.6837 & 0.5762 & 0.698 & 0.7738 & 0.6311\\
FRA-space & FRA4 &  0.6868 & 0.5748 & 0.6956 & 0.7748 & 0.6324\\
FRA-space* & FRA4 & 0.6874 & 0.5833 & 0.6994 & 0.7872 & 0.6505\\
DEU-time & FRA4 &  0.6986 & 0.5848 & 0.7053 & 0.7778 & 0.6364\\
DEU0 & FRA4 & 0.6792 & 0.577 & 0.693 & 0.7783 & 0.6388 \\
FRA4$^{\dagger}$ & FRA4 & 0.6591 & 0.5522 & 0.675 & 0.7566 & 0.6086 \\

\hline
- &	FRA5 & 0.6027 & 0.4871 & 0.6055 & 0.7194 & 0.5611 \\
FRA-base & FRA5 & 0.6457 & 0.5440 & 0.6692 & 0.7574 & 0.6096\\
FRA-time & FRA5 & 0.6540 &	0.5490 & 0.6694 & 0.7916 & 0.6154\\
FRA-time* & FRA5 & 0.6584 & 0.5508 & 0.6747 & 0.7645 & 0.6188\\
FRA-space & FRA5 &  0.6456 & 0.5386 & 0.6617 & 0.7582 & 0.6101\\
FRA-space* & FRA5 & 0.6511 & 0.557 & 0.682 & 0.7602 & 0.615\\
DEU-time & FRA5 &  0.6512 & 0.5500 & 0.6726 & 0.7600 & 0.6230\\
DEU0 & FRA5 & 0.6458 & 0.5407 & 0.6604 & 0.7561 & 0.6094 \\
FRA5$^{\dagger}$ & FRA5 & 0.6232 & 0.4981 & 0.6214 & 0.732 & 0.5799\\

\hline
- &	FRA6 & 0.5701 & 0.4651 & 0.5924 & 0.7098 & 0.5501\\
FRA-base & FRA6 & 0.6402 & 0.5230 & 0.6497 & 0.7507 & 0.6009\\
FRA-time & FRA6 & 0.6269 & 0.5198 & 0.6439 & 0.7591 & 0.6117 \\
FRA-time* & FRA6 & 0.6296 & 0.5255 & 0.6554 & 0.7408 & 0.5882\\
FRA-space & FRA6 & 0.6311 & 0.5196 & 0.6460 & 0.7311 & 0.5762 \\
FRA-space* & FRA6 & 0.6278 & 0.5228 & 0.647 & 0.7233 & 0.6005\\
DEU-time & FRA6 & 0.6283 & 0.5302 & 0.6540 & 0.7545 & 0.6058\\
DEU0 & FRA6 & 0.6218 & 0.5209 & 0.6406 & 0.7444 & 0.5942 \\
FRA6$^{\dagger}$ & FRA6 & 0.5646 & 0.4677 & 0.5921 & 0.7121 & 0.5577\\

\hline
- &	FRA7 &  0.5360 & 0.4087 & 0.5354 & 0.6704 & 0.5227\\
FRA-base & FRA7 & 0.6038 & 0.4954 & 0.6253 & 0.7226 & 0.5656\\
FRA-time & FRA7 & 0.6061 & 0.5017 & 0.6280 & 0.7406 & 0.5881 \\
FRA-time** & FRA7 & 0.6150 & 0.4990 & 0.6265 & 0.7351 & 0.5779\\
FRA-space & FRA7 & 0.5966 & 0.4874 & 0.6144 & 0.7210 & 0.5637 \\
FRA-space** & FRA7 & 0.6268 & 0.5113 & 0.6393 & 0.7367 & 0.5831\\
DEU-time & FRA7 & 0.6036 & 0.4895 & 0.6138 & 0.7255 & 0.5693\\
DEU0 & FRA7 & 0.5658 & 0.4719 & 0.5868 & 0.7269 & 0.5721 \\
FRA7$^{\dagger}$ & FRA7 & 0.5249 & 0.4193 & 0.5411 & 0.6936 & 0.531 \\

\hline
\end{tabular}
\end{center}
\caption{Results in France. FRA-time** and FRA-space** pre-train datasets denote training with $201$k size queue. * supervised pre-training, $\dagger$ CSCL pre-training \cite{cscl}}
\label{FRA_results}
\end{table}

\begin{table}[!h]
\begin{center}
\begin{tabular}{|c|c|c|c|c|c|c|c|c|}
\hline
\multicolumn{2}{|c|}{datasets} & \multicolumn{3}{|c|}{macro} & \multicolumn{2}{|c|}{micro}\\
\hline
pre-train & train & acc & iou & F1 & acc & iou\\
\hline \hline
- &	GHA0 &  0.5232 & 0.3737 & 0.5313 & 0.5977 & 0.4262 \\
GHA-base &	GHA0 & 0.5429 & 0.3923 & 0.5509 & 0.6441 & 0.4431 \\
SSU-base &	GHA0 & 0.5305 & 0.3832 & 0.5427 & 0.6152 & 0.4380\\
SSU0 &	GHA0 & 0.5373 & 0.3840 & 0.5466 & 0.6381 & 0.4401\\
GHA0$^{\dagger}$ & GHA0 & 0.5228 & 0.3804 & 0.5363 & 0.6026 & 0.4312\\

\hline
- &	GHA1 & 0.4554 & 0.3198 & 0.4663 & 0.5506 & 0.3799 \\
GHA-base &	GHA1 & 0.4960 & 0.3435 & 0.4958 & 0.5652 & 0.3939 \\
SSU-base &	GHA1 & 0.4765 & 0.3371 & 0.4881 & 0.5620 & 0.3919\\
SSU0 &	GHA1 & 0.4932 & 0.3367 & 0.4862 & 0.5596 & 0.3882\\
GHA1$^{\dagger}$ & GHA1 & 0.4558 & 0.3265 & 0.4674 & 0.5542 & 0.3883\\

\hline
- &	GHA2 & 0.4208 & 0.2809 & 0.4260 & 0.5065 & 0.3391 \\
GHA-base &	GHA2 & 0.4383 & 0.2989 & 0.4469 & 0.5362 & 0.3663 \\
SSU-base &	GHA2 & 0.4422 & 0.3069 & 0.4514 & 0.5437 & 0.3733\\
SSU0 &	GHA2 & 0.4357 & 0.2950 & 0.4406 & 0.5246 & 0.3556\\
GHA2$^{\dagger}$ & GHA2 & 0.4367 & 0.2937 & 0.4388 & 0.525 & 0.3559\\

\hline
- &	GHA3 & 0.3845 & 0.2502 & 0.3917 & 0.4795 & 0.3153 \\
GHA-base &	GHA3 & 0.4338 & 0.2777 & 0.4277 & 0.4908 & 0.3365 \\
SSU-base &	GHA3 & 0.4398 & 0.2821 & 0.4273 & 0.5126 & 0.3446\\
SSU0 &	GHA3 & 0.4008 & 0.2674 & 0.4105 & 0.5003 & 0.3336\\
GHA3$^{\dagger}$ & GHA3 & 0.3978 & 0.2622 & 0.4074 & 0.4742 & 0.3108\\

\hline
- &	GHA4 & 0.3517 & 0.2230 & 0.3514 & 0.4482 & 0.2880
 \\
GHA-base &	GHA4 & 0.3917 & 0.2590 & 0.3985 & 0.4976 & 0.3312 \\
SSU-base &	GHA4 & 0.3880 & 0.2474 & 0.3887 & 0.4930 & 0.3252\\
SSU0 &	GHA4 & 0.3780 & 0.2383 & 0.3736 & 0.4536 & 0.2933\\
GHA4$^{\dagger}$ & GHA4 & 0.353 & 0.2266 & 0.3560 & 0.4513 & 0.2914\\

\hline
- &	GHA5 & 0.3594 & 0.2257 & 0.3577 & 0.4402 & 0.2822 \\
GHA-base &	GHA5 & 0.3784 & 0.2375 & 0.3646 & 0.4606 & 0.2992 \\
SSU-base &	GHA5 & 0.3836 & 0.2405 & 0.3673 & 0.4674 & 0.3005\\
SSU0 &	GHA5 & 0.3611 & 0.2330 & 0.3622 & 0.4733 & 0.3100\\
GHA5$^{\dagger}$ & GHA5 & 0.3522 & 0.2223 & 0.3512 & 0.4420 & 0.2855\\

\hline
\end{tabular}
\end{center}
\caption{Results in Ghana. * supervised pre-training, $\dagger$ CSCL pre-training \cite{cscl}}
\label{GHA_results}
\end{table}

\begin{table}[!h]
\begin{center}
\begin{tabular}{|c|c|c|c|c|c|c|c|c|}
\hline
\multicolumn{2}{|c|}{datasets} & \multicolumn{3}{|c|}{macro} & \multicolumn{2}{|c|}{micro}\\
\hline
pre-train & train & acc & iou & F1 & acc & iou\\
\hline \hline
- &	SSU0 & 0.5767 & 0.4333 & 0.5979 & 0.6289 & 0.4587\\
SSU-base &	SSU0 & 0.621 & 0.4764 & 0.6395 & 0.6719 & 0.5059\\
SSU-time &	SSU0 & 0.6338 & 0.4881 & 0.6502 & 0.6775 & 0.5180\\
% SSU-space &	SSU0 & 0.6232 & 0.4794 & 0.6429 & 0.674 & 0.5083\\
GHA-base &	SSU0 & 0.6296 & 0.4863 & 0.6493 & 0.6785 & 0.5139\\
SSU0$^{\dagger}$ & SSU0 & 0.6056 & 0.4542 & 0.6197 & 0.6573 & 0.4895\\

\hline
- &	SSU1 & 0.5009 & 0.3593 & 0.5199 & 0.5728 & 0.4013\\
SSU-base &	SSU1 & 0.5348 & 0.3923 & 0.5585 & 0.5992 & 0.4305\\
SSU-time &	SSU1 & 0.5467 & 0.4016 & 0.5656 & 0.6109 & 0.4398\\
% SSU-space &	SSU1 & 0.5360 & 0.3962 & 0.5605 & 0.6023 & 0.4319\\
GHA-base &	SSU1 & 0.5401 & 0.4000 & 0.5627 & 0.6064 & 0.4352\\
GHA0 &	SSU1 & 0.4947 & 0.3527 & 0.5140 & 0.5657 & 0.3963\\
SSU1$^{\dagger}$ & SSU1 & 0.5290 & 0.3837 & 0.5463 & 0.5881 & 0.4166 \\

\hline
- &	SSU2 & 0.4232 & 0.2898 & 0.4403 & 0.5271 & 0.3579\\
SSU-base &	SSU2 & 0.4478 & 0.311 & 0.4663 & 0.5420 & 0.3718\\
SSU-time &	SSU2 & 0.4629 & 0.3308 & 0.4883 & 0.5500 & 0.3793\\
% SSU-space &	SSU2 & 0.4522 & 0.319 & 0.4730 & 0.5459 & 0.3744\\
GHA-base &	SSU2 & 0.4675 & 0.3290 & 0.4871 & 0.5529 & 0.3821\\
GHA0 &	SSU2 & 0.4225 & 0.2897 & 0.4401 & 0.5244 & 0.3554\\
SSU2$^{\dagger}$ & SSU2 & 0.4454 & 0.3083 & 0.4620 & 0.5406 & 0.3704\\

\hline
- &	SSU3 & 0.3914 & 0.2573 & 0.3984 & 0.4984 & 0.3319\\
SSU-base &	SSU3 & 0.4176 & 0.2841 & 0.4338 & 0.5023 & 0.3354\\
SSU-time &	SSU3 & 0.4265 & 0.2901 & 0.4389 & 0.5285 & 0.3592\\
% SSU-space &	SSU3 & 0.4233 & 0.2866 & 0.4359 & 0.5101 & 0.3412\\
GHA-base &	SSU3 & 0.4008 & 0.2793 & 0.4269 & 0.5140 & 0.3459\\
GHA0 &	SSU3 & 0.3806 & 0.2544 & 0.3957 & 0.4793 & 0.3152\\
SSU3$^{\dagger}$ & SSU3 & 0.3904 & 0.2607 & 0.4035 & 0.4910 & 0.3254\\

\hline
- &	SSU4 & 0.3317 & 0.2108 & 0.3334 & 0.4536 & 0.2933\\
SSU-base &	SSU4 & 0.3855 & 0.2601 & 0.4021 & 0.4952 & 0.3303\\
SSU-time &	SSU4 & 0.3855 & 0.2601 & 0.4021 & 0.4972 & 0.3318\\
% SSU-space &	SSU4 & 0.3832 & 0.256 & 0.3971 & 0.4945 & 0.3290\\
GHA-base &	SSU4 & 0.3803 & 0.2518 & 0.3915 & 0.4939 & 0.3280\\
GHA0 &	SSU4 & 0.3278 & 0.2104 & 0.3297 & 0.4490 & 0.2895\\
SSU4$^{\dagger}$ & SSU4 & 0.3467 & 0.2232 & 0.3534 & 0.4525 & 0.2959\\

\hline
- &	SSU5 & 0.2625 & 0.1649 & 0.2719 & 0.3802 & 0.2347\\
SSU-base &	SSU5 & 0.3052 & 0.1986 & 0.3216 & 0.4255 & 0.2702\\
SSU-time &	SSU5 & 0.3258 & 0.2133 & 0.3432 & 0.4255 & 0.2702\\
% SSU-space &	SSU5 & 0.3072 & 0.2011 & 0.3277 & 0.4229 & 0.2682\\
GHA-base &	SSU5 & 0.3180 & 0.2059 & 0.3314 & 0.4240 & 0.2690\\
GHA0 &	SSU5 & 0.2798 & 0.1819 & 0.3007 & 0.3831 & 0.2369\\
SSU5$^{\dagger}$ & SSU5 & 0.2847 & 0.1804 & 0.2949 & 0.3879 & 0.2406\\

\hline
\end{tabular}
\end{center}
\caption{Results in S.Sudan. * supervised pre-training, $\dagger$ CSCL pre-training \cite{cscl}}
\label{SSU_results}
\end{table}

\clearpage

\clearpage

\ifCLASSOPTIONcaptionsoff
  \newpage
\fi

{\small
\bibliographystyle{IEEEtran}
\bibliography{ms}
}
% \end{IEEEbiography}


\end{document}